\makeatletter
\def\input@path{{package/}}
\makeatother

\documentclass{egpubl}
\usepackage{eg2026}

\ConferencePaper     
\CGFStandardLicense

\usepackage[T1]{fontenc}
\usepackage{package/dfadobe}  

\usepackage{cite}  
\BibtexOrBiblatex
\electronicVersion
\PrintedOrElectronic
\ifpdf \usepackage[pdftex]{graphicx} \pdfcompresslevel=9
\else \usepackage[dvips]{graphicx} \fi

\usepackage{egweblnk} 
\usepackage{lineno}

\usepackage{amsmath}
\usepackage{amssymb}
\usepackage{booktabs}
\usepackage{multirow}
\usepackage{xcolor}
\usepackage{makecell}
\usepackage{placeins}
\usepackage[normalem]{ulem}

\newcommand{\ie}{i.\,e.\ }	

\newcommand{\revised}[1]{\textcolor{black}{#1}}

\title{HiMat: DiT-based Ultra-High Resolution SVBRDF Generation}

\author[Z. Wang et al.]
{\parbox{\textwidth}{\centering Zixiong Wang$^{1}$\orcid{0000-0002-6170-7339}, Jian Yang$^{1}$\orcid{0000-0003-4800-832X}, Yiwei Hu$^{2}$\orcid{0000-0002-3674-295X}, Milo\v{s} Ha\v{s}an$^{2,3}$\orcid{0000-0003-3808-6092}, Beibei Wang\thanks{Corresponding author}$^{4}$\orcid{0000-0001-8943-8364}}
        \\
{\parbox{\textwidth}{\centering $^1$College of Computer Science, Nankai University,
         $^2$Adobe Research,
         $^3$NVIDIA,
         $^4$Nanjing University
       }
}
}

\begin{document}

\teaser{
 \includegraphics[width=0.86\linewidth]{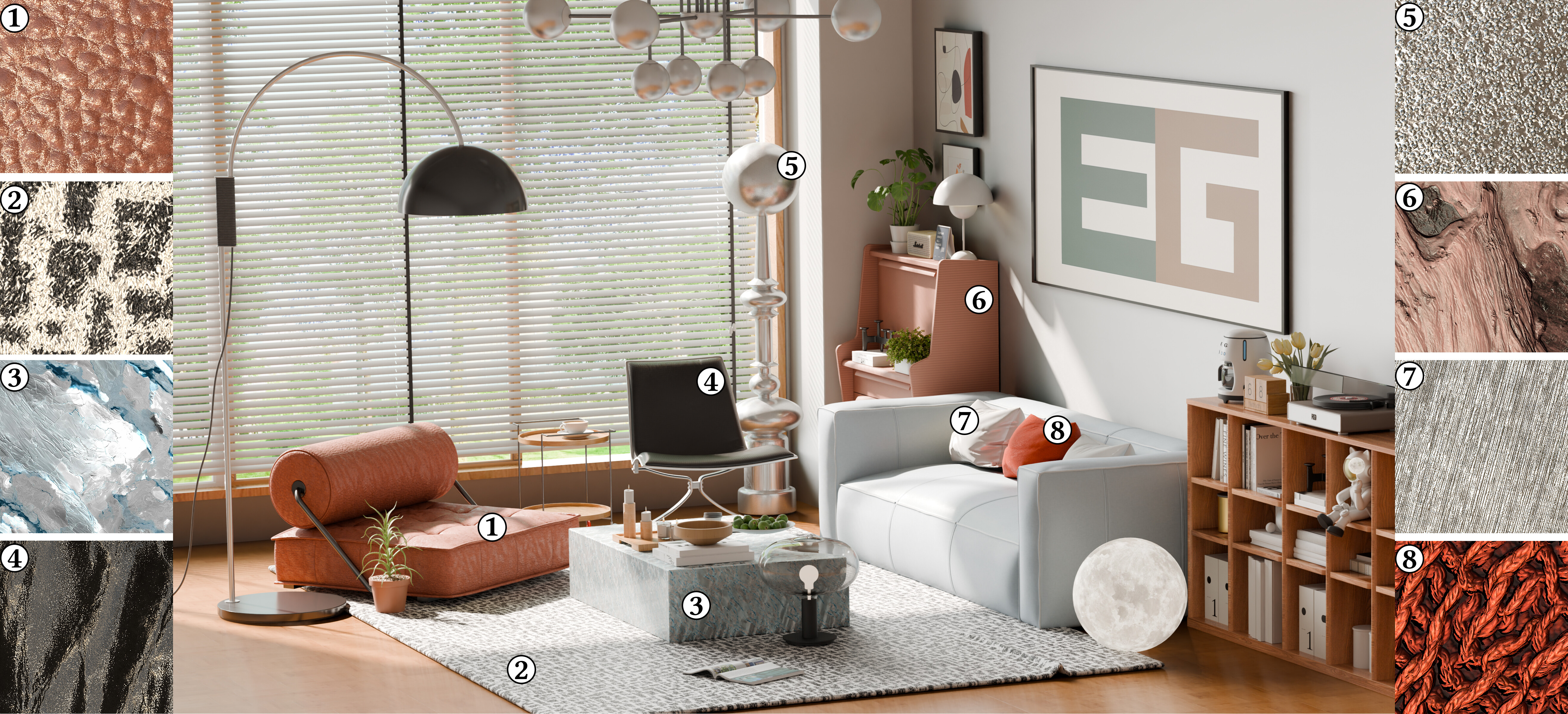}
 \centering
 \caption{We present HiMat, a diffusion-based framework generating ultra-high-resolution ($4096 \times 4096$) SVBRDF materials from text prompts. Our approach achieves this resolution while preserving high-frequency details crucial for meso-structure components such as normal and height maps. We showcase a variety of materials within a single scene, highlighting the preserved fine-scale details and texture fidelity.}
\label{fig:teaser}
}

\maketitle
\begin{abstract}
Creating ultra-high-resolution spatially varying bidirectional reflectance functions (SVBRDFs) is critical for photorealistic 3D content creation, to faithfully represent fine-scale surface details required for close-up rendering. However, achieving 4K generation faces two key challenges: (1) the need to synthesize multiple reflectance maps at full resolution, which multiplies the pixel budget and imposes prohibitive memory and computational cost, and (2) the requirement to maintain strong pixel-level alignment across maps at 4K, which is particularly difficult when adapting pretrained models designed for the RGB image domain. We introduce HiMat, a diffusion-based framework tailored for efficient and diverse 4K SVBRDF generation. To address the first challenge, HiMat generates in a high-compression latent space via a DC-AE and employs a pretrained diffusion transformer with linear attention to improve per-map efficiency. To address the second challenge, we propose CrossStitch, a lightweight convolutional module that enforces cross-map consistency without incurring the cost of global attention. Our experiments show that HiMat achieves high-fidelity 4K SVBRDF generation with superior efficiency, structural consistency, and diversity compared to prior methods. Beyond materials, our framework also generalizes to related applications such as intrinsic decomposition.


\end{abstract}  

\begin{figure*}[tp]
    \centering
    \includegraphics[width=\linewidth]{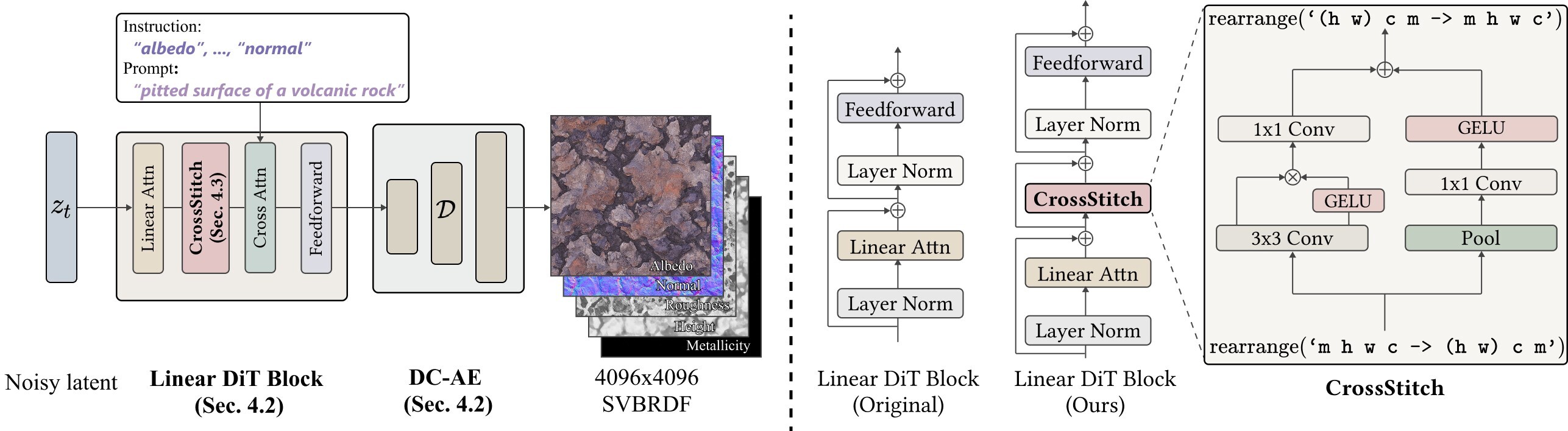}

    \caption{\textbf{Overview.} 
    \textbf{Left}: Given text instructions, our framework generates 4K SVBRDF maps through a latent denoising pipeline based on linear DiT (Sec.~\ref{sssec:dit}), with outputs reconstructed by a deep compression autoencoder (DC-AE) (Sec.~\ref{sssec:dcae}). \emph{CrossStitch} layers (Sec.~\ref{ssec:CrossStitch}) are integrated into the linear DiT block after each linear attention layer.
    The combination of linear DiT and DC-AE enables efficient ultra-high-resolution generation, while the CrossStitch design ensures consistency across maps.
    \textbf{Right}: Architecture of our modified DiT block (cross-attention omitted for clarity). A lightweight convolutional \emph{CrossStitch} module enables localized feature exchange across maps, ensuring pixel alignment.}
    \label{fig:pipeline}
\end{figure*}
\section{Introduction}
\label{sec:intro}
Modeling the reflectance properties of spatially-varying bidirectional reflectance functions (SVBRDFs) is a fundamental task in photorealistic rendering. To reduce production effort and the need for heavy manual intervention, automatic generation has emerged as a promising alternative. For such methods to achieve high-quality results, two requirements are particularly critical: ultra-high resolution (\ie, 4K) and material diversity. High resolution enables fine-grained detail in close-up views. At the same time, diversity ensures broad coverage of real-world materials and supports scalable asset creation for applications such as games, visual effects, and architectural and product visualization.

Achieving both goals, however, remains challenging. First, 4K generation imposes extreme demands on GPU memory and computational throughput, limiting practical scalability. Furthermore, the limited scale of existing accessible SVBRDF datasets~\cite{deschaintre2018single, ma2023opensvbrdf, vecchio2023matsynth} prevents generative models from capturing the vast appearance space of real-world materials, resulting in limited diversity and poor generalization.

While recent generative approaches have made progress, they still fall short of meeting these demands. Most existing techniques are based on generative adversarial networks (GANs)~\cite{Guo:2020:MaterialGAN, tilegen, zhou2023PhotoMat} or diffusion-based methods~\cite{he23text2mat, vecchio2024matfuse, xue2024reflectancefusion, xin2024dreampbr}. These typically operate at low resolutions (e.g., $512\times512$) and rely on limited synthetic datasets, which restrict both quality and diversity. MatGen~\cite{vecchio2024controlmat} pioneers 4K material generation with a cascaded denoising pipeline~\cite{du2024demofusion}, but suffers from low efficiency and error accumulation. To improve diversity, MaterialPicker~\cite{ma2024materialpicker} fine-tunes pretrained video diffusion models to generate material maps rather than RGB frames, using softmax attention~\cite{attentionall17}. While such fine-tuning achieves diversity from the pretrained prior, the quadratic attention cost limits the input resolution to $256 \times 256$, making the approach impractical for 4K generation in the near future.

In this paper, we propose \emph{HiMat}, a novel framework for efficient and diverse 4K SVBRDF generation. We identify two key challenges in 4K SVBRDF generation: (1) each of multiple reflectance maps must be generated at full 4K resolution, multiplying the pixel budget and causing prohibitive memory and compute cost, and (2) these maps must remain strictly pixel-aligned at 4K, a requirement that is particularly difficult when adapting pretrained image models designed for 3-channel RGB inputs.

To tackle the first issue, our core idea is to reduce the effective pixel budget and improve per-map processing efficiency. We achieve this by performing generation in a high-compression latent space and replacing quadratic attention with a more efficient alternative. Specifically, we leverage a deep compression autoencoder (DC-AE)\cite{chen2025dcae} to compress 4K inputs while preserving key reflectance properties, and a linear diffusion transformer\cite{xie2025sana} to accelerate per-map generation at ultra-high resolution.
For the second issue, our key insight is that SVBRDF maps, being inherently pixel-aligned and limited in number, allow consistency to be enforced without resorting to costly global attention. Based on this, we design \emph{CrossStitch}, a lightweight convolution-based module enabling alignment across maps. Convolution is hardware-friendly and benefits from mature optimizations such as Winograd~\cite{lavin2016fast}, making CrossStitch both efficient and scalable. This allows us to adapt pretrained image diffusion models to the SVBRDF domain, effectively exploiting strong priors while preserving inter-map consistency. Notably, CrossStitch is non-destructive and remains compatible with a wide range of latent diffusion architectures (e.g., U-Net).

Our results demonstrate that HiMat enables high-fidelity 4K SVBRDF generation with enhanced visual quality and reduced computational cost (see Fig.~\ref{fig:teaser}). Thanks to its lightweight design, HiMat produces $4096 \times 4096$ SVBRDFs in 90 seconds on consumer-grade hardware such as an NVIDIA RTX 4090D (in 20 steps). Further, our framework can be generalized to other tasks, including intrinsic decomposition. Our main contributions are:
\begin{itemize}
    \item A memory-efficient and computationally scalable diffusion model for native ultra-high-resolution SVBRDF generation,
    \item A lightweight CrossStitch module that captures localized inter-map dependencies, enabling structural consistency across SVBRDF maps, and can be non-destructively added to latent diffusion models,
\end{itemize}

\section{Related Work}
\label{sec:related}
\revised{
Material creation and editing have a long history in graphics. Early efforts explored the use of text~\cite{memery2023generating} and images~\cite{DADDB19} to enable more intuitive authoring and control. Below, we focus on generative model-based approaches that aim to automate material generation through deep learning techniques. For a comprehensive overview of artistic authoring and editing techniques across appearance, we refer the readers to surveys~\cite{schmidt2016state, kavoosighafi2024deep}.
}
\subsection{Material Generation}
Learning-based material generation has attracted increasing attention for simplifying material creation. Early works leveraged GANs~\cite{Guo:2020:MaterialGAN, tilegen} trained on synthetic datasets. To improve realism, PhotoMat~\cite{zhou2023PhotoMat} utilizes real materials captured with flash photographs. \revised{Based on the generator of PhotoMat, DiffMat~\cite{DiffMat24} introduces an auxiliary diffusion network to enhance the latent representation of flash photographs, leading to improved reconstruction quality.} Meanwhile, transformer-based approaches have been explored for procedural material generation~\cite{guerrero2022matformer, hu2023gen}, aiming to enhance scalability and diversity. However, due to the inherent limitations of GANs, these methods are difficult to scale to 4K SVBRDF generation.

Recently, diffusion-based methods have emerged as a promising direction for improving SVBRDF generation. \revised{Text2Mat~\cite{he23text2mat}, DreamPBR~\cite{xin2024dreampbr}, and ReflectanceFusion~\cite{xue2024reflectancefusion} adopt a dual-phase design for richer semantics: the first stage generates a latent representation of a natural image using pretrained image models, and the second stage recovers SVBRDF parameters from the retrained decoder. While this strategy enhances diversity, retraining the decoder on a limited SVBRDF dataset limits overall reconstruction quality. Moreover, DreamPBR employs a post-hoc super-resolution module, which often leads to over-smoothed outputs and missing some fine-grain details.}
In contrast, some methods aim to train models to generate SVBRDF parameters directly. Among these, MatFuse and MatGen~\cite{vecchio2024matfuse, vecchio2024controlmat} are trained from scratch in the style of latent diffusion models (LDMs)\cite{rombach2022LDM}, with multiple encoders conditioned on text, image, or sketch inputs. However, due to limited training data, these models often struggle with diversity and generalization. MaterialPicker\cite{ma2024materialpicker}, on the other hand, fine-tunes a softmax-attention video diffusion model to generate SVBRDFs while retaining pretrained priors. Still, it is currently limited to a $256 \times 256$ resolution and requires a post hoc super-resolution module to upscale the outputs.

\subsection{Diffusion Models Architectures}
Diffusion models are generative frameworks for producing diverse, high-quality images~\cite{yang2023diffusion}, with architectures evolving from convolutional U-Nets augmented by self-attention~\cite{dhariwal2021diffusion, ramesh2022hierarchicaltextconditionalimagegeneration, rombach2022LDM}. Recently, Diffusion Transformers (DiT-s)\cite{Peebles_2023DiT} have emerged, leveraging pure attention-based architectures\cite{attentionall17}. Stable Diffusion 3~\cite{esser2024sd3}, Flux~\cite{flux2024}, Lumina-Image~\cite{lumina2}, and many other recent models extend this paradigm by improving quality, efficiency, and cross-modal alignment.

However, the quadratic complexity of softmax attention limits the scalability of DiT models, particularly for high-resolution image and video generation~\cite{flux2024, sora24, yang2024cogvideox}. This has motivated the development of more efficient alternatives, such as linear attention~\cite{xie2025sana} and state-space models~\cite{phung2024dimsum, hu2024zigma}. In this work, we adopt a linear attention-based Diffusion Transformer, Sana \cite{xie2025sana,2025sana15} for 4K SVBRDF generation, achieving high visual fidelity with significantly reduced memory and computational costs.

\subsection{Ultra-High Resolution Diffusion Models}
Generating 4K RGB images is crucial for many applications but remains computationally demanding. Existing approaches can be broadly divided into two categories: training-free cascade pipelines and model-based architectural adaptations.

Training-free cascade methods~\cite{du2024demofusion, kim2025diffusehigh} progressively upscale diffusion outputs without retraining, using techniques such as residual connections and low-frequency injection. However, their multi-stage design introduces error accumulation and incurs high latency (e.g., DiffuseHigh~\cite{kim2025diffusehigh} takes 258s per 4K image on an H100 GPU), making them impractical for generating multiple correlated SVBRDF maps.

In contrast, model-based strategies adapt diffusion architectures for 4K generation. PixArt-$\Sigma$\cite{chen2024pixart} and Sana\cite{xie2025sana} redesign transformers for improved efficiency, while Diffusion4K~\cite{zhang2025diffusion4k} combines high-compression VAEs with wavelet supervision but applies uniform weights across all frequency bands. URAE~\cite{yu2025urae} further provides practical guidelines for ultra-resolution training. \revised{Unlike standard image generation, SVBRDF generation requires the simultaneous generation of multiple consistent maps. To this end, we adopt a model-based fine-tuning strategy and introduce a learnable CrossStitch module to enforce cross-map alignment during generation effectively.}
\section{Preliminaries}
\label{sec:prelimi}
\paragraph*{Latent Diffusion Models.} Latent diffusion models (LDMs)\cite{rombach2022LDM} perform denoising in a compressed latent space, enabling efficient high-resolution synthesis. \revised{Specifically, a variational autoencoder (VAE)\cite{Kingma2014VAE} encodes an input RGB image $\boldsymbol{x} \in \mathbb{R}^{H \times W \times 3}$ into a latent representation $\boldsymbol{z}_0 = \mathcal{E}(\boldsymbol{x}) \in \mathbb{R}^{\hat{H} \times \hat{W} \times C}$, where $\mathcal{E}$ denotes the encoder. Here, $H$ and $W$ are the spatial dimensions of the input image, $C$ is the number of latent channels, and $\hat{H} = \tfrac{H}{F}$ and $\hat{W} = \tfrac{W}{F}$ represent the height and width of the latent representation, respectively, downsampled by a factor of $F$.}

The forward process corrupts latent $\boldsymbol{z}_0$ by adding Gaussian noise:
\begin{equation}
\boldsymbol{z}_t = \alpha_t \cdot \boldsymbol{z}_0 + \sigma_t \cdot \epsilon, \quad \epsilon \sim \mathcal{N}(0, 1),
\end{equation}
with $\alpha_t, \sigma_t$ controlling the noise schedule.

In LDM~\cite{rombach2022LDM}, the denoising network $\Theta$ predicts the noise.
Most recent variants (e.g. SD3~\cite{esser2024sd3}) use velocity prediction via flow matching~\cite{lipman2023flow}:
\begin{equation}
v_{\Theta}(\boldsymbol{z}_t, t, \boldsymbol{c}_\text{text}) = \epsilon - \boldsymbol{z}_0.
\label{eq:reflow}
\end{equation}

The network $\Theta$ is typically a U-Net or DiT with self-attention for global context and cross-attention for conditioning.

\paragraph*{SVBRDF.} An SVBRDF typically represents the parameters of the Cook-Torrance microfacet model~\cite{cook1982reflectance} with a GGX distribution function~\cite{walter2007microfacet}. In our setting, an SVBRDF of size $H \times W$ consists of albedo $\boldsymbol{a}\in \mathbb{R}^{H \times W \times 3}$, normal $\boldsymbol{n}\in \mathbb{R}^{H \times W \times 3}$, roughness $\boldsymbol{r}\in \mathbb{R}^{H \times W}$, metallicity $\boldsymbol{m}\in \mathbb{R}^{H \times W}$, and height $\boldsymbol{h}\in\mathbb{R}^{H \times W}$. This yields a set of maps:
\begin{equation}
\boldsymbol{I} = \left\{ \boldsymbol{a}, \boldsymbol{n}, \boldsymbol{r}, \boldsymbol{m}, \boldsymbol{h} \right\},
\quad \boldsymbol{I} \in \mathbb{R}^{M \times H \times W \times 3}.
\end{equation}
For efficient processing, we concatenate the scalar maps $\boldsymbol{r}$, $\boldsymbol{m}$, and $\boldsymbol{h}$ into a single three-channel image, resulting in $M=3$ maps.

\section{Method}
\label{sec:method}

\subsection{Overview}
Our goal is to train a diffusion-based SVBRDF generator that produces 4K materials from text prompts. High-quality generation requires efficiency, diversity, and consistency across reflectance maps. The key challenges are twofold: (1) each map must be generated at full 4K resolution, and the presence of multiple maps multiplies the pixel budget, leading to prohibitive memory and computational cost, and (2) the maps are physically interdependent and must remain pixel-aligned at 4K, a requirement that is particularly difficult when adapting pretrained image models designed initially for 3-channel RGB inputs rather than multi-channel SVBRDFs.

To address these challenges, we propose \emph{HiMat}, a framework for efficient, diverse, and consistent 4K SVBRDF generation (see Fig.~\ref{fig:pipeline}). To overcome the first challenge, HiMat employs a high-compression autoencoder and a linear-attention diffusion transformer (Sec.~\ref{sssec:dcae}) to reduce the effective pixel budget for 4K SVBRDF, substantially lowering memory consumption and computational cost. To address the second issue, HiMat introduces the CrossStitch module (Sec.~\ref{ssec:CrossStitch}), a lightweight convolution-based design that efficiently enforces pixel-level alignment across maps. In addition, to improve data quality and enhance material diversity, we incorporate prompt-based dataset augmentation with enriched textual descriptions (Sec.~\ref{ssec:text}).

\begin{figure}[t]
    \centering
    \includegraphics[width=\linewidth]{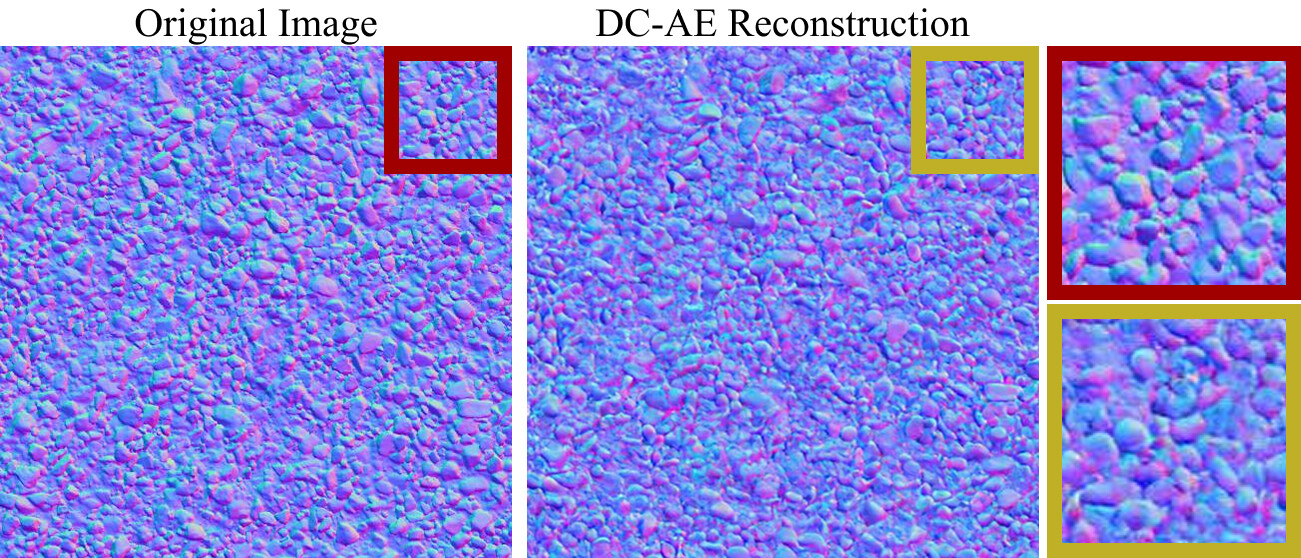}
    \caption{Normal map reconstruction quality with DC-AE. 
    Color bias in the reconstructed normals indicates distribution mismatch with ground truth, motivating our fine-tuning of the decoder for SVBRDF maps.}
    \label{fig:vae_show}
\end{figure}

\subsection{A Lightweight Latent Diffusion Model for 4K SVBRDFs}

\label{ssec:issue_1}
To mitigate the prohibitive pixel budgets of generating multiple 4K maps, HiMat integrates two complementary designs: a Deep Compression AutoEncoder (DC-AE) to reduce the effective pixel count, and a Linear-Attention Diffusion Transformer to accelerate per-map processing.

\paragraph*{High-Compression Autoencoder.}
\label{sssec:dcae}

High-resolution latent diffusion requires compact yet expressive representations. However, standard VAEs with $F{=}8$ compression~\cite{rombach2022LDM} incur excessive overhead at 4K, often leading to OOM errors.

To address this, we adopt DC-AE~\cite{chen2025dcae}, which achieves up to $F{=}32$ compression via residual encoding and resolution-aware adaptation. This reduces 4K inputs to $128 \times 128$ latent features, enabling tractable diffusion without compromising reconstruction fidelity.

However, since DC-AE is originally trained on natural image datasets, it may not fully capture the unique characteristics of SVBRDF maps, particularly for normal maps that require accurate preservation of geometric details and orientation information (see Fig.~\ref{fig:vae_show}). To better align DC-AE with the SVBRDF domain, we follow latent diffusion models~\cite{rombach2022LDM} and fine-tune the decoder using a combination of pixel-wise loss $\mathcal{L}_{\mathrm{rec}}$ and perceptual loss $\mathcal{L}_{\mathrm{LPIPS}}$~\cite{zhang2018perceptual}:
\begin{equation}
\label{eq::vae_loss}
\mathcal{L}_{\mathrm{vae}} = \lambda_{\mathrm{rec}} \mathcal{L}_{\mathrm{rec}} + \lambda_{\mathrm{LPIPS}} \mathcal{L}_{\mathrm{LPIPS}}.
\end{equation}
We omit adversarial loss due to its instability at ultra-high resolutions~\cite{chen2025dcae}. This adaptation yields a compact latent space while faithfully preserving SVBRDF-specific details essential for high-fidelity 4K generation.


\paragraph*{Efficient Diffusion Transformers.}
\label{sssec:dit}
DiT~\cite{Peebles_2023DiT} adopts stacked softmax attention blocks~\cite{attentionall17} for global context modelling. Given an input sequence $\boldsymbol{s} \in \mathbb{R}^{N \times C}$, where $N$ denotes the sequence length and $C$ the channel dimension, the softmax attention is defined as:
\begin{equation}
\text{Attention}({Q}, {K}, {V}) = \text{Softmax}\left( \frac{{Q} {K}^\top}{\sqrt{d_k}} \right) {V},
\end{equation}
where ${Q}, {K}, {V}$ are the query, key, and value matrices obtained via learned linear projections.  

While effective, softmax attention incurs $\mathcal{O}(N^2)$ complexity, making it impractical for ultra-high resolution image and video generation~\cite{sora24,yang2024cogvideox}, as evidenced by the 256$\times$256 resolution cap in MaterialPicker~\cite{ma2024materialpicker}.

To improve scalability, HiMat adopts a linear attention-based DiT~\cite{xie2025sana,2025sana15}, which reduces complexity to $\mathcal{O}(N)$:
\begin{equation}
\text{LinearAttention}({Q}, {K}, {V}) = 
\frac{\operatorname{ReLU}(Q)\left(\operatorname{ReLU}(K)^\top V\right)}
{\operatorname{ReLU}(Q)\left(\operatorname{ReLU}(K)^\top \mathbf{1}\right)},
\end{equation}
where $\mathbf{1}$ is an all-ones vector. This design preserves global receptive fields while significantly lowering memory footprint and runtime cost, enabling efficient scaling to 4K.

Together, DC-AE and linear-attention DiT directly alleviate the pixel burden, making high-resolution SVBRDF generation both tractable and efficient.




\begin{figure}[t]
    \centering
    \includegraphics[width=\linewidth]{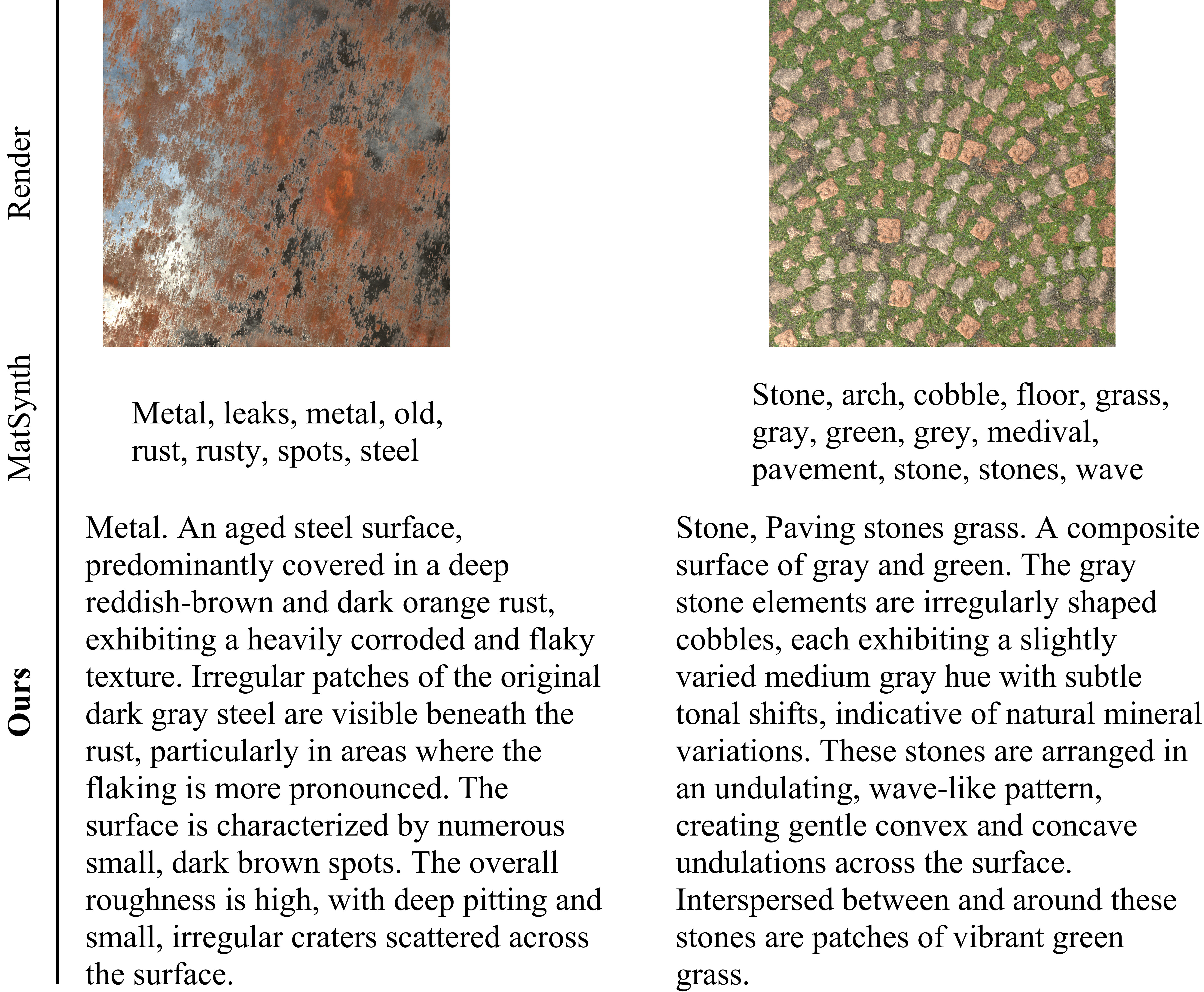}
    \caption{Textual description comparison. MatSynth~\cite{vecchio2023matsynth} provides short keyword labels, whereas our method generates rich, perceptually faithful descriptions that better capture material appearance and structure.}
    \label{fig:text_train}
\end{figure}
\subsection{CrossStitch: Enforcing Cross-Map Consistency in 4K SVBRDFs}
\label{ssec:CrossStitch}

While the previous designs address the prohibitive cost of processing individual 4K maps, a second challenge is enforcing consistency across multiple reflectance maps. Since SVBRDF channels are physically coupled, even minor misalignments at 4K resolution can introduce shading discontinuities and implausible reflectance.

A natural idea is to adapt video-generation strategies, where softmax-attention layers enforce coherence across frames~\cite{ma2024materialpicker}. However, SVBRDF maps differ fundamentally from video: they are few in number, inherently pixel-aligned, and free of temporal drift. Consequently, softmax attention is both inefficient (quadratic cost) and unnecessary. Linear attention, although reducing complexity to linear in very long sequences, offers no advantage in this setting, as the small number of maps leaves the complexity effectively quadratic.

Our key insight is that SVBRDF maps only require local neighborhood communication rather than global attention. To this end, we design a \emph{CrossStitch} module, a lightweight convolution-based layer that sparsely exchanges information across maps while preserving spatial alignment. Convolution is hardware-friendly and benefits from mature optimisations such as Winograd~\cite{lavin2016fast}, making CrossStitch scalable to ultra-high resolutions.

Formally, given latent features $\boldsymbol{f} \in \mathbb{R}^{M \times \hat{H} \times \hat{W} \times \hat{C}}$ after self-attention, where $M$ is the number of SVBRDF maps and $\hat{C}$ the channel dimension, CrossStitch rearranges $\boldsymbol{f}$ to align the map dimension as channels, applies a 1D convolution across maps, and restores the original layout (using einops~\cite{rogozhnikov2022einops}):
\begin{equation}
\begin{aligned}
\boldsymbol{f} &\leftarrow \mathrm{rearrange}(\boldsymbol{f}, \texttt{`m h w c -> (h w) c m'}) \\
\boldsymbol{f} &\leftarrow \mathrm{CrossStitch}(\boldsymbol{f}) \\
\boldsymbol{f} &\leftarrow \mathrm{rearrange}(\boldsymbol{f}, \texttt{`(h w) c m -> m h w c'}).
\end{aligned}
\end{equation}

In practice, \emph{CrossStitch} is a dual-branch 1D convolutional module. One branch applies a depthwise-separable convolution (spatial $3\times3$ followed by pointwise $1\times1$) for efficient local feature mixing, while the other aggregates information across maps via average pooling, followed by a $1\times1$ convolution and GELU activation to capture shared semantic context. \revised{To integrate this module into the pretrained diffusion network, we insert it after the self-attention layer.} All convolutional layers are zero-initialized and connected via residual connections to preserve pretrained representations. Although numerous convolutional variants exist~\cite{li2021convsurvey}, our simple yet effective design achieves both local integration and global alignment without incurring the overhead of self-attention.

CrossStitch is non-destructive, maintaining the structural integrity of each map while enforcing semantic and spatial coherence across them. This enables us to efficiently adapt pretrained image generation models to the material domain and leverage their learned priors for 4K SVBRDF generation.

\begin{figure*}[t]
    \centering
    \includegraphics[width=\linewidth]{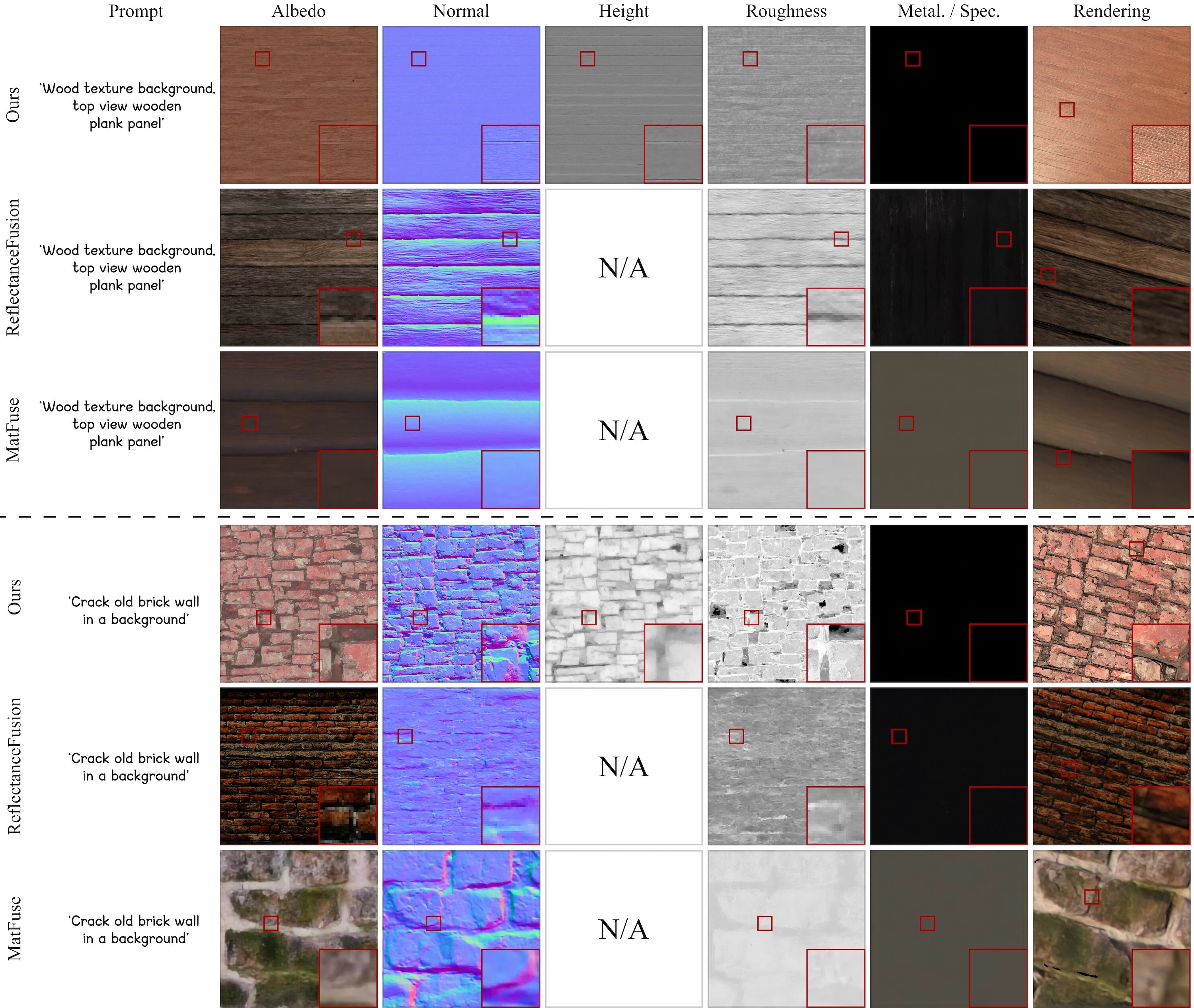}
    \caption{\revised{Visual comparison between HiMat, ReflectanceFusion~\cite{xue2024reflectancefusion}, and MatFuse~\cite{matfusion}. ReflectanceFusion exhibits baked-in lighting artifacts and is limited to a resolution of $256\times256$. MatFuse suffers from reduced realism and diversity due to training exclusively on synthetic data at $512\times512$ resolution. In contrast, HiMat delivers high-quality 4K materials with fine detail. A slightly tilted camera view is employed in the rendering to visualize the details better.}
    }  
    \label{fig:mat_compare}
\end{figure*}

\subsection{Enhancing Material Diversity with Richer Text Prompts}
\label{ssec:text}
While efficiency and consistency address computational bottlenecks, the diversity of generated materials remains constrained by the sparsity of supervision in existing datasets. Public SVBRDF datasets such as MatSynth~\cite{vecchio2023matsynth} and the dataset proposed by Deschaintre et al.~\cite{deschaintre2018single} provide only tag-style labels, whose descriptive power is insufficient to guide generative models effectively.

To overcome this limitation, we introduced prompt enrichment during both training and inference. During training, we expand the original tag annotations using large language models (LLMs) with carefully designed templates (detailed in the supplementary materials), \revised{inspired by DTDMat~\cite{chen2024dtdmat}}. These templates explicitly capture intrinsic material attributes, such as color, texture, roughness, and surface imperfections, yielding more descriptive and diverse prompts that align better with generative objectives (Fig.~\ref{fig:text_train}).

At inference, we leverage a lightweight local LLM (Gemma-2~\cite{team2024gemma}) as the text encoder and reuse the same templates to generate system-level prompts that serve as semantic guides. This design draws inspiration from Lumina-Image-2.0~\cite{lumina2}, which shows that system prompts can significantly improve generation quality without requiring architectural changes. By enriching supervision at both stages, our approach expands the diversity of generated materials and improves generalization to unseen categories, complementing the efficiency and consistency modules described earlier.
\begin{table}[b]
\centering
\caption{Quantitative comparison for SVBRDF generation with MatFuse~\cite{vecchio2024matfuse} and ReflectanceFusion~\cite{xue2024reflectancefusion}.}
\label{tab:svbrdf}
\resizebox{\linewidth}{!}{%
\begin{tabular}{c|ccc} 
\toprule
\multicolumn{1}{l|}{}                      & MatFuse & ReflectanceFusion & Ours  \\ 
\midrule
CLIPScore $\uparrow$                       & 25.91   &  30.22            & \textbf{30.27}     \\ 
\midrule
Q-Align (quality) $\uparrow$                         & 2.17    & 2.34              & \textbf{3.23}     \\
Q-Align (aesthetic) $\uparrow$                         & 1.63    & 1.98              & \textbf{2.33}     \\
Aesthetics $\uparrow$                      & 3.87    & 4.13             & \textbf{4.63}     \\ 
\midrule
\multicolumn{1}{c|}{GLCM Score $\uparrow$} & 0.63    & 0.31              & \textbf{0.96}     \\
\bottomrule
\end{tabular}
}
\end{table}

\begin{figure*}[t]
    \centering
    \includegraphics[width=\linewidth]{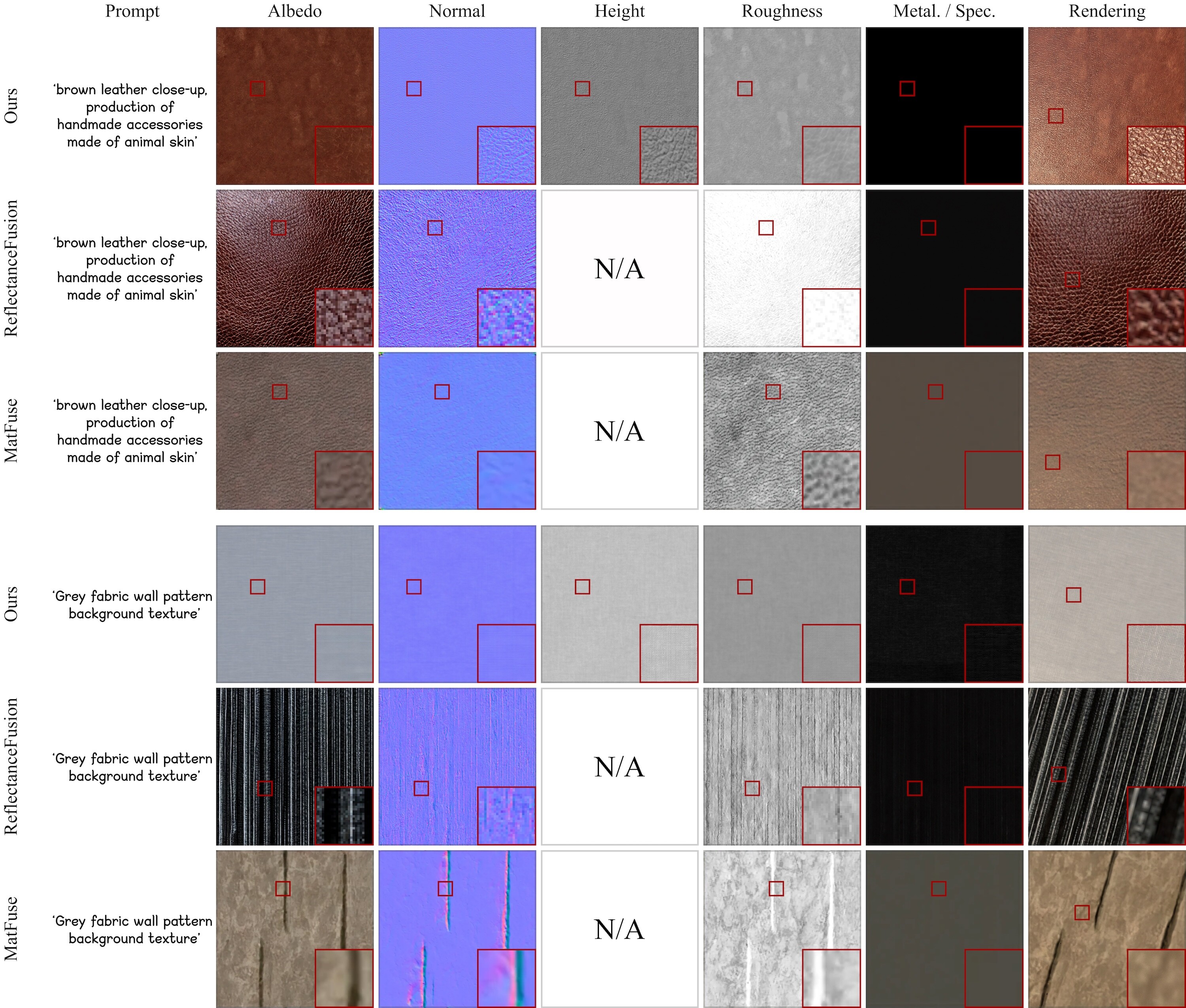}
    \caption{\revised{Further visual comparison between HiMat, ReflectanceFusion~\cite{xue2024reflectancefusion} and MatFuse~\cite{vecchio2024matfuse}. 
    Note the baked-in lighting artifact at the top of the albedo map in the second row of the ReflectanceFusion result.}}
    \label{fig:mat_compare2}
\end{figure*}

\begin{figure*}[t]
    \centering
    \includegraphics[width=\linewidth]{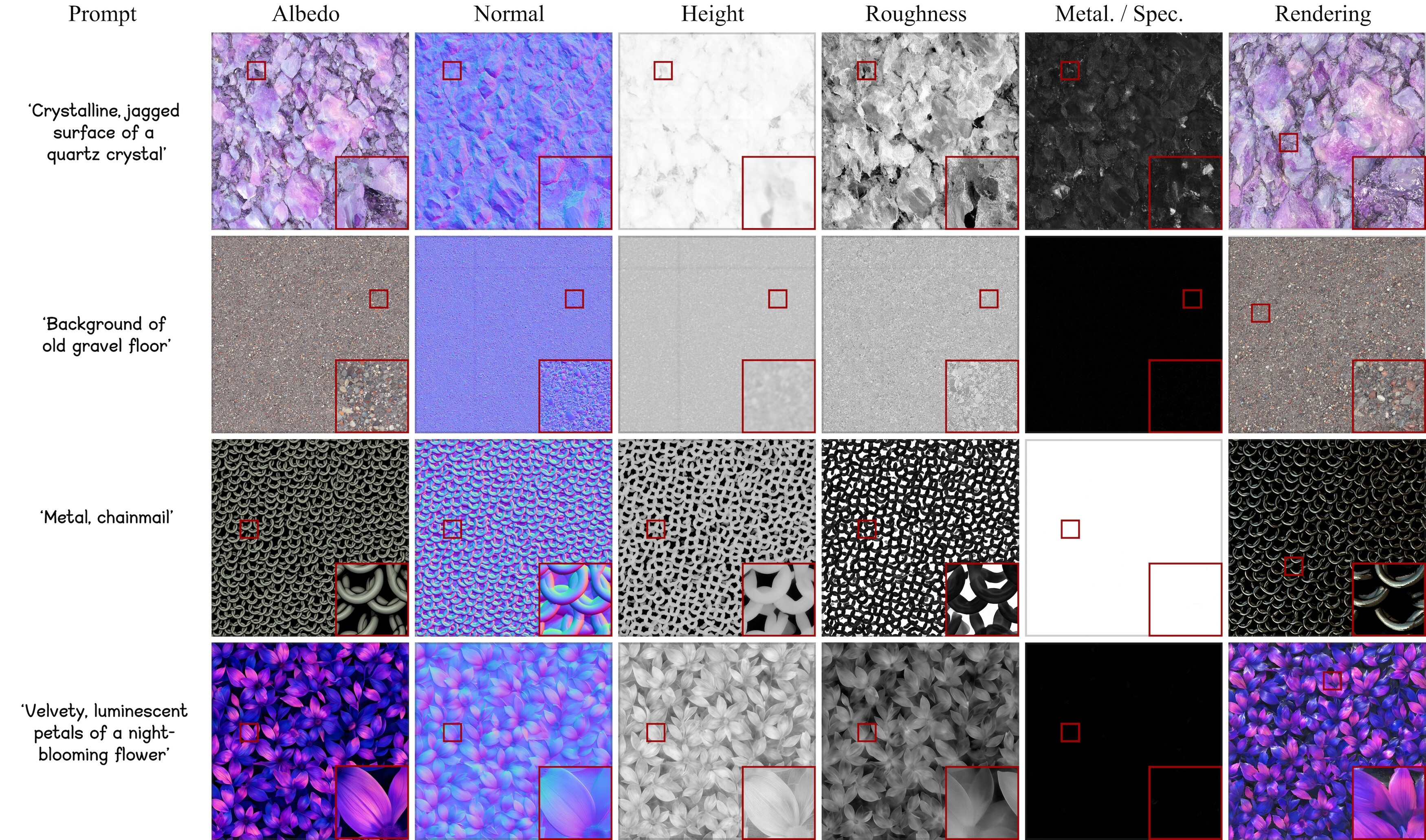}
    \caption{\revised{Additional visual results from HiMat. These examples further demonstrate the diversity, realism, and fine structural details achieved by our method.}}
    \label{fig:mat_vis}
\end{figure*}

\section{Results}

\begin{figure}[t]
    \centering
    \includegraphics[width=\linewidth]{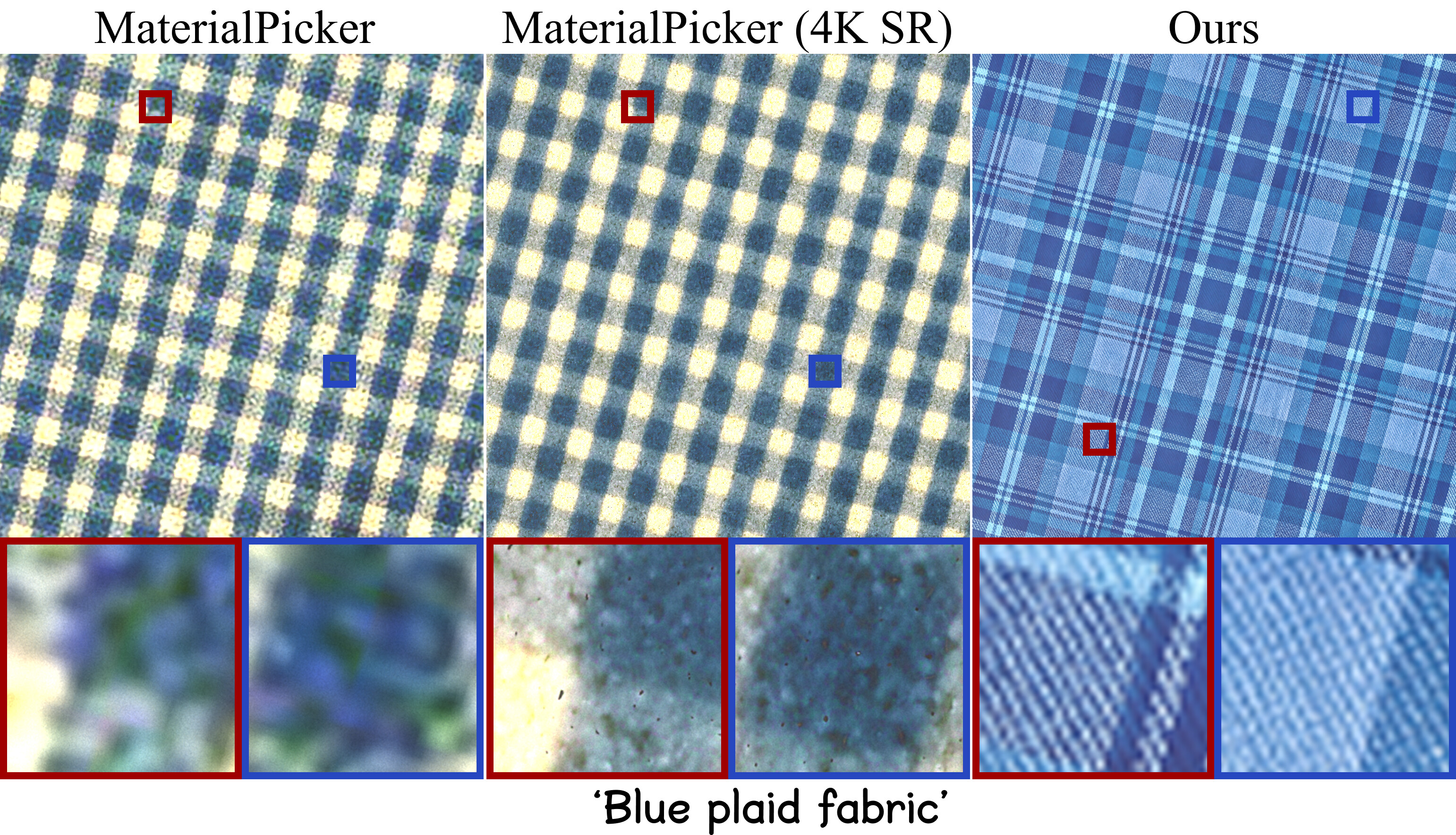}
    \caption{\revised{Visual comparison between our native 4K SVBRDF generation and the outputs of MaterialPicker, including its upscaled 4K version using SUPIR~\cite{yu2024scaling}. Recovering fine fabric microstructures from low-resolution MaterialPicker outputs is challenging, underscoring the advantages of directly generating high-resolution results.}
    }
    \label{fig:4kSR}
\end{figure}
\subsection{Implementation Details}
We trained our model on the combined datasets of MatSynth~\cite{vecchio2023matsynth} and the dataset proposed by Deschaintre et al.~\shortcite{deschaintre2018single}, for a total of 6,198 unique physically-based rendering~(PBR) materials. The text prompts are augmented with Gemini 2.5~\cite{comanici2025gemini}. To address the issue of uneven distribution for material types~(e.g., Wood and Metal), we resample different categories to the same large number. For training, we initialize our model from the pre-trained Sana-1024px checkpoint~\cite{xie2025sana} with 1.6B parameters. To effectively capture high-frequency material details while maintaining training stability, we employ a progressive-resolution strategy, gradually increasing the training resolution from $1024 \times 1024$ to $2048 \times 2048$ and finally to $4096 \times 4096$. To support tileable material generation, we additionally adopt noise rolling~\cite{vecchio2024controlmat}.

\subsection{Metrics}
Since ground-truth data is unavailable for the generation task, we follow the evaluation protocol of MatFuse~\cite{vecchio2024matfuse}, which assesses SVBRDF quality through rendered images. Specifically, we render the generated SVBRDFs under environment lighting at $4096 \times 4096$ and evaluate them across three aspects: semantic alignment measured by CLIPScore~\cite{hessel2021clipscore}, perceptual image quality measured by Q-Align~\cite{qalign} (quality / aesthetic), respectively, and aesthetics score~\cite{schuhmann2022laion}, and high-frequency detail preservation measured by the Gray Level Co-occurrence Matrix (GLCM)~\cite{zhang2025diffusion4k}. Higher values indicate better performance for all metrics. The complete set of text prompts is provided in the supplemental materials.

\subsection{Text Conditioned Generation}
We compare our method against MatFuse~\cite{vecchio2024matfuse} and ReflectanceFusion~\cite{xue2024reflectancefusion} on 500 text prompts (adopted from ReflectanceFusion, detailed in the supplementary material). ReflectanceFusion is limited to $256 \times 256$ and MatFuse to $512 \times 512$, whereas our approach natively supports 4K. As shown in Fig.~\ref{fig:mat_compare} and~\ref{fig:mat_compare2}, both baselines exhibit low-resolution artifacts that fail under close-up inspection. MatFuse produces less plausible and less diverse materials due to its exclusive reliance on synthetic training data, missing the strong priors learned from real images that both ReflectanceFusion and our approach exploit. ReflectanceFusion, by contrast, achieves relatively higher quality but suffers from baked-lighting artifacts since it relies on a base text-to-image model that generates RGB images with highlights and shadows, which are difficult to remove during its second-stage recovery (see the first column of Fig.~\ref{fig:mat_compare}). In comparison, HiMat generates diverse, realistic, and detailed 4K SVBRDFs that preserve fidelity even under close-up views.

In Tab.~\ref{tab:svbrdf}, we present quantitative comparisons across multiple metrics. MatFuse obtains the lowest CLIPScore and Q-Align scores, reflecting limited plausibility and diversity. ReflectanceFusion achieves a higher CLIPScore (30.22) owing to its base text-to-image backbone; however, its reliance on a $256 \times 256$ resolution yields weak structural fidelity, as evidenced by its low GLCM score. In contrast, HiMat achieves the best performance across nearly all metrics, demonstrating not only greater material diversity but also superior preservation of fine structural detail. Additional qualitative results are shown in Fig.~\ref{fig:mat_vis}.




\paragraph*{MaterialPicker vs. Ours.}
\revised{
We compare our method against MaterialPicker~\cite{ma2024materialpicker}, which leverages video diffusion priors by fine-tuning a pretrained model for material generation. Since its output is limited to $256 \times 256$, we apply SUPIR~\cite{yu2024scaling} to upscale each map to $4096 \times 4096$. As shown in Fig.~\ref{fig:4kSR}, our native 4K generation preserves fine details significantly better, whereas recovering fabric microstructures from the low-resolution outputs of MaterialPicker remains challenging. This highlights the advantages of directly generating high-resolution results.
}

\begin{figure}[htp]
    \centering
    \includegraphics[width=\linewidth]{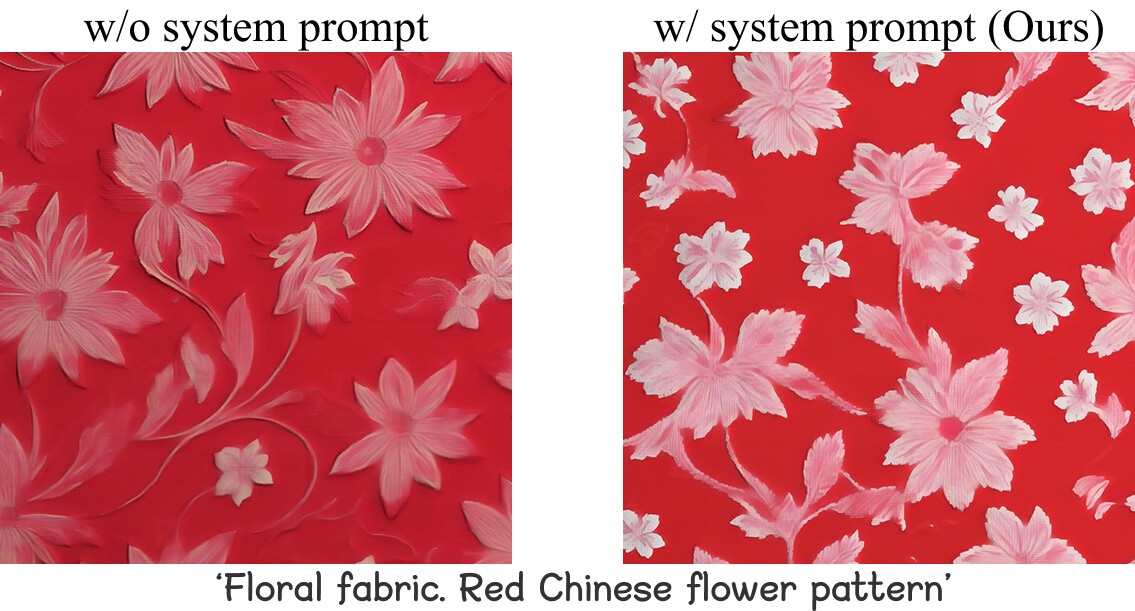}
    \caption{Comparison of generations with and without system-level prompts. Incorporating our designed templates yields finer structure and detail.}
    \label{fig:abl_chi}
\end{figure}

\begin{figure}[htp]
    \centering
    \includegraphics[width=\linewidth]{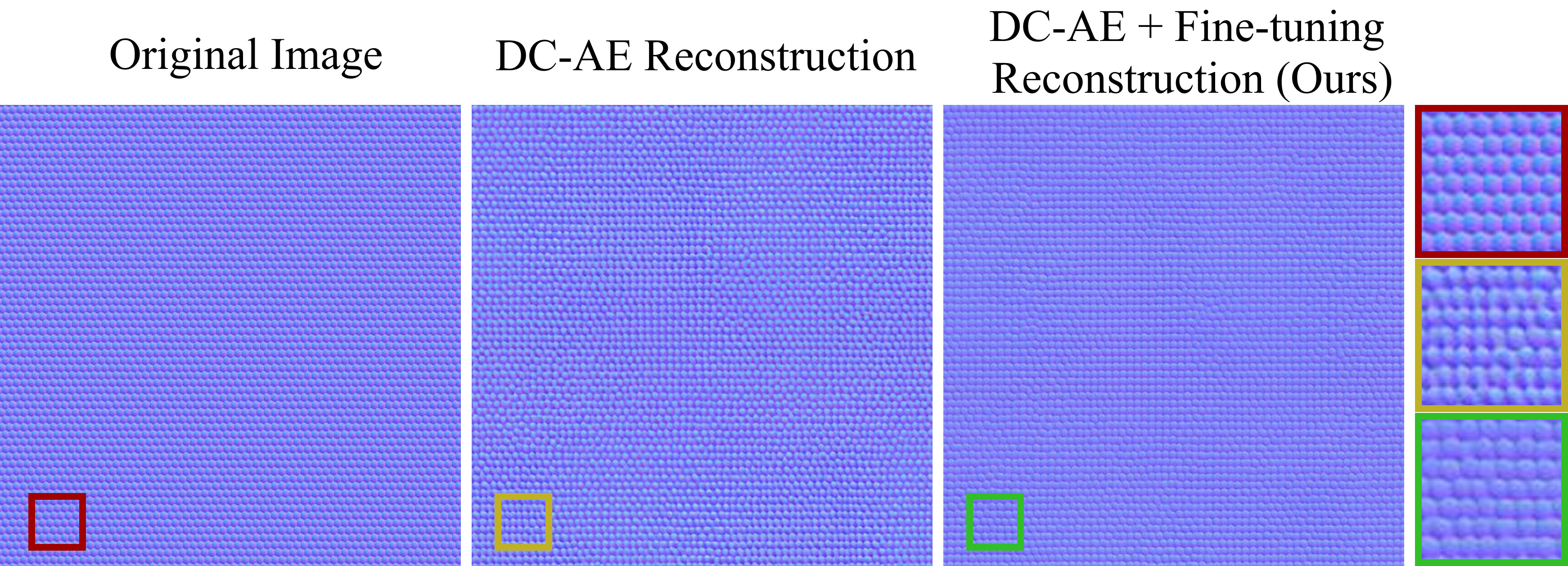}
    \caption{Effect of DC-AE fine-tuning. Fine-tuned models produce normal maps with improved spatial structure and orientations. Zoom in for better visualization.}
    \label{fig:abl_vae}
\end{figure}

\subsection{Ablation studies and Analysis}
\paragraph*{Textual Prompt Enrichment.} To evaluate the effectiveness of textual prompt enrichment, we render materials from the original dataset and compare the original tag-based descriptions with our enhanced long-form prompts using CLIP-Score~\cite{hessel2021clipscore}. While MatSynth achieves a CLIP-Score of 27.74, our enriched prompts yield a score of 29.26, indicating improved textual-image alignment through simple yet targeted expansions. Furthermore, we assess the impact of system-level prompts during inference by comparing generations with and without our designed template. As shown in Fig.~\ref{fig:abl_chi}, incorporating system-level prompts yields richer texture detail without additional efforts. 

\paragraph*{VAE Fine-Tuning.}  
To assess the benefits of fine-tuning the DC-AE decoder, we compare models before and after fine-tuning. As reported in Tab.~\ref{tab:vae_finetune}, our fine-tuned variant consistently improves reconstruction quality across multiple metrics, including Fréchet Inception Distance (rFID), Peak Signal-to-Noise Ratio (PSNR), Root Mean Square Error (RMSE), Structural Similarity Index Measure (SSIM), and Learned Perceptual Image Patch Similarity (LPIPS). Visual comparisons in Fig.\ref{fig:abl_vae} further confirm fine-tuning enhanced structural fidelity on 4K MatSynth~\cite{vecchio2023matsynth} results.

\begin{table}[t]
\centering
\caption{Quantitative evaluation of decoder fine-tuning for DC-AE on the MatSynth~\cite{vecchio2023matsynth} dataset at 4K resolution.}
\label{tab:vae_finetune}
\resizebox{\linewidth}{!}{%
\begin{tabular}{c|ccccc} 
\toprule
Model          & rFID $\downarrow$  & RMSE $\downarrow$  & PSNR $\uparrow$      & SSIM $\uparrow$    & LPIPS $\downarrow$  \\ 
\midrule
DC-AE          & 1.50               & 0.10               & 28.71                & 0.75               & 0.16                \\
DC-AE + fine-tuning & $\mathbf{1.29}$ & $\mathbf{0.08}$ & $\mathbf{30.28}$ & $\mathbf{0.79}$ & $\mathbf{0.13}$  \\
\bottomrule
\end{tabular}
}
\end{table} 

\begin{figure}[htp]
    \centering
    \includegraphics[width=\linewidth]{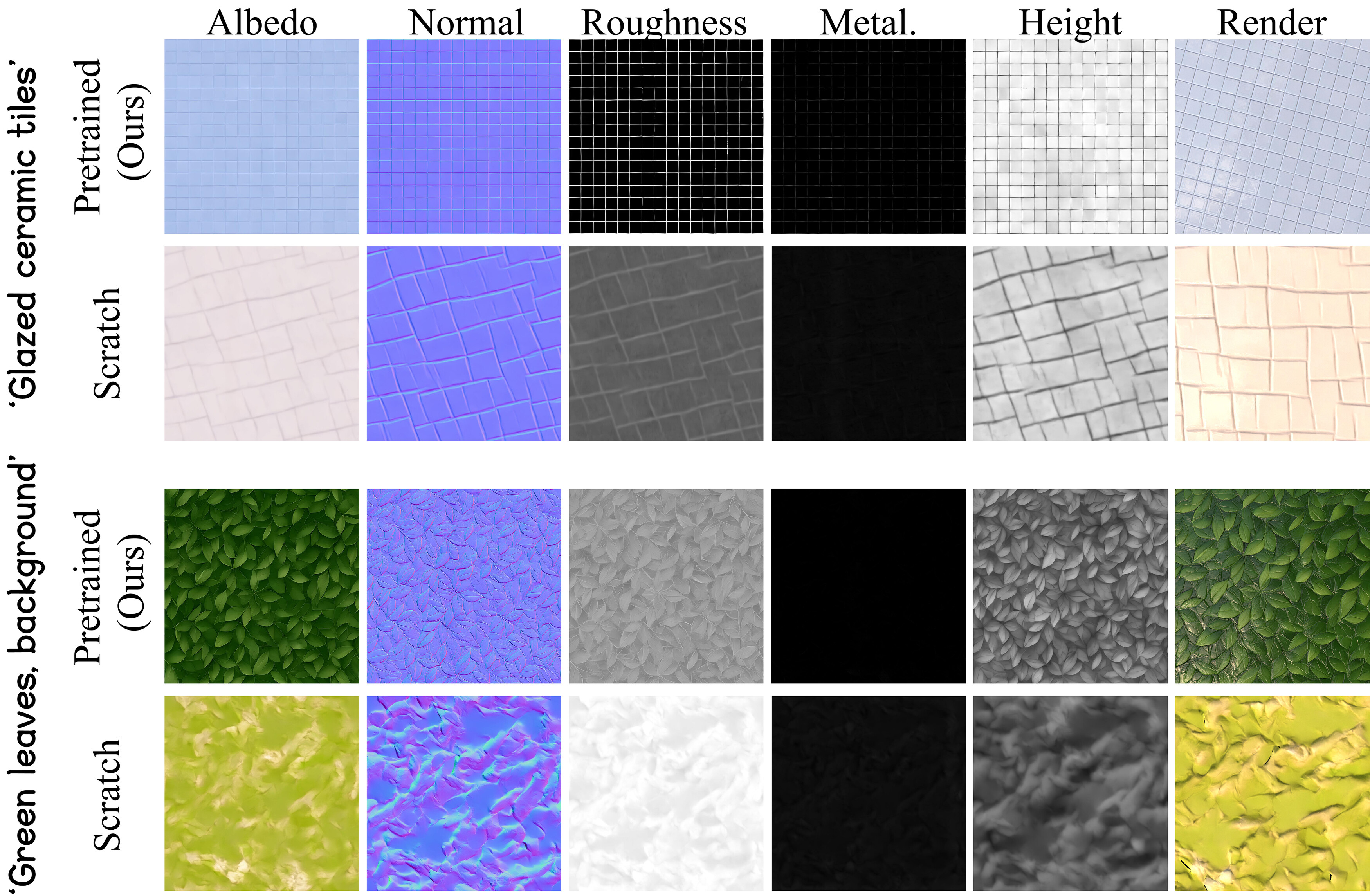}
    \caption{Effect of pretrained priors. Models trained from scratch exhibit limited diversity, whereas pretrained initialization enables higher-quality and more diverse material generation.}
    \label{fig:abl_pretrain}
\end{figure}


\paragraph*{Pretraining vs. Training from Scratch.}  
To validate the impact of pretrained prior, we compare two $1024 \times 1024$ models with identical architectures: one trained from scratch and the other initialized from our pretrained weights. As shown in Fig.~\ref{fig:abl_pretrain}, the scratch model collapses to categories in the trainset and fails to generate diverse types such as `leaves'. In contrast, the pretrained model achieves higher quality and better diversity, highlighting the importance of strong priors.

\begin{table*}[t]
\centering
\caption{
Comparison of computational cost during inference. All results are measured on a consumer-level RTX 4090D GPU.
We report both absolute results and relative ratios (normalized by CrossStitch variant, denoted as 1.00) for variants with attention and linear attention.
}
\label{tab:flops}
\resizebox{\linewidth}{!}{%
\begin{tabular}{l|c|ccc|ccc|ccc} 
\toprule
                 & \multirow{2}{*}{Params (B)} & \multicolumn{3}{c|}{Forward FLOPs (T)$\downarrow$}                                & \multicolumn{3}{c|}{Memory (GB)$\downarrow$}                                      & \multicolumn{3}{c}{Time (s/step)$\downarrow$}                                   \\ 
\cmidrule(l){3-11}
                 &                             & 1024×1024            & 2048×2048             & 4096×4096              & 1024×1024             & 2048×2048             & 4096×4096             & 1024×1024            & 2048×2048            & 4096×4096             \\ 
\midrule
Attention        & 2.01 / 1.14                 & 11.20 / 1.21         & 43.68 / 1.22          & 173.58 / 1.22          & 10.89 / 1.05          & 12.76 / 1.01          & 20.44 / 1.03          & 0.39 / 1.30          & 1.23 / 1.29          & 5.13 / 1.28           \\
Lin. ttn.        & 2.01 / 1.14                 & 11.22 / 1.21         & 43.75 / 1.22          & 173.88 / 1.22          & 11.59 / 1.12          & 15.80 / 1.25          & OOM                   & 0.43 / 1.43          & 1.30 / 1.37          & -                     \\
Ours (CrossStitch) & \textbf{1.76 / 1.00}        & \textbf{9.25 / 1.00} & \textbf{35.87 / 1.00} & \textbf{142.36 / 1.00} & \textbf{10.39 / 1.00} & \textbf{12.63 / 1.00} & \textbf{19.93 / 1.00} & \textbf{0.30 / 1.00} & \textbf{0.95 / 1.00} & \textbf{4.01 / 1.00}  \\
\bottomrule
\end{tabular}
}
\end{table*}

\paragraph*{CrossStitch vs. Attention.}
We evaluate our CrossStitch-based architecture against standard softmax attention and linear attention baselines. We first analyze performance across resolutions, then compare training dynamics by training all three variants at $1024 \times 1024$ under identical settings. In Tab.~\ref{tab:flops}, we replace CrossStitch with either standard or linear attention and report parameter count, forward FLOPs, peak memory usage, and inference time per step, all measured on an RTX 4090D GPU. Our CrossStitch model consistently reduces computational overhead: at $1024 \times 1024$ and $2048 \times 2048$, it achieves up to 22\% fewer FLOPs and 25\% less memory than linear attention, while also running faster. At $4096 \times 4096$, both standard and linear attention incur prohibitive costs, with linear attention leading to out-of-memory failures. 

\begin{table}[t]
\centering
\caption{\revised{Quantitative comparison of intrinsic decomposition performance on the Hypersim~\cite{roberts2021hypersim} test dataset. Our retrained results demonstrate that CrossStitch delivers consistent improvements under identical data conditions. Note that our method does not match the original RGB$\leftrightarrow$X results, as the original model was trained on a large-scale dataset.
}}
\label{tab:abl_rgbx}
\resizebox{.8\linewidth}{!}{%
\begin{tabular}{c|c|cc} 
\toprule
                                                            &            & \multicolumn{2}{c}{RGB$\leftrightarrow$X (retrained)}  \\
                                                            &            & w/ CS          & w/o CS                                \\ 
\midrule
\multirow{3}{*}{PSNR$\uparrow$}                             & Albedo     & \textbf{13.16} & 12.40                                 \\
                                                            & Normal     & \textbf{15.07} & 13.83                                 \\
                                                            & Irradiance & \textbf{16.42} & 15.94                                 \\
                                                            & Mean       & \textbf{14.89} & 14.05                                 \\ 
\midrule
\multirow{3}{*}{LPIPS$\downarrow$}                          & Albedo     & \textbf{0.43}  & 0.47                                  \\
                                                            & Normal     & \textbf{0.40}  & 0.48                                  \\
                                                            & Irradiance & \textbf{0.37}  & 0.41                                  \\
                                                            & Mean       & \textbf{0.40}  & 0.45                                  \\ 
\midrule
\multicolumn{1}{l|}{\multirow{4}{*}{DreamSim $\downarrow$}} & Albedo     & \textbf{0.21}  & 0.24                                  \\
\multicolumn{1}{l|}{}                                       & Normal     & \textbf{0.13}  & 0.17                                  \\
\multicolumn{1}{l|}{}                                       & Irradiance & \textbf{0.18}  & 0.21                                  \\
\multicolumn{1}{l|}{}                                       & Mean       & \textbf{0.17}  & 0.21                                  \\
\bottomrule
\end{tabular}
}
\end{table}

\begin{figure}[t]
    \centering  
    \includegraphics[width=\linewidth]{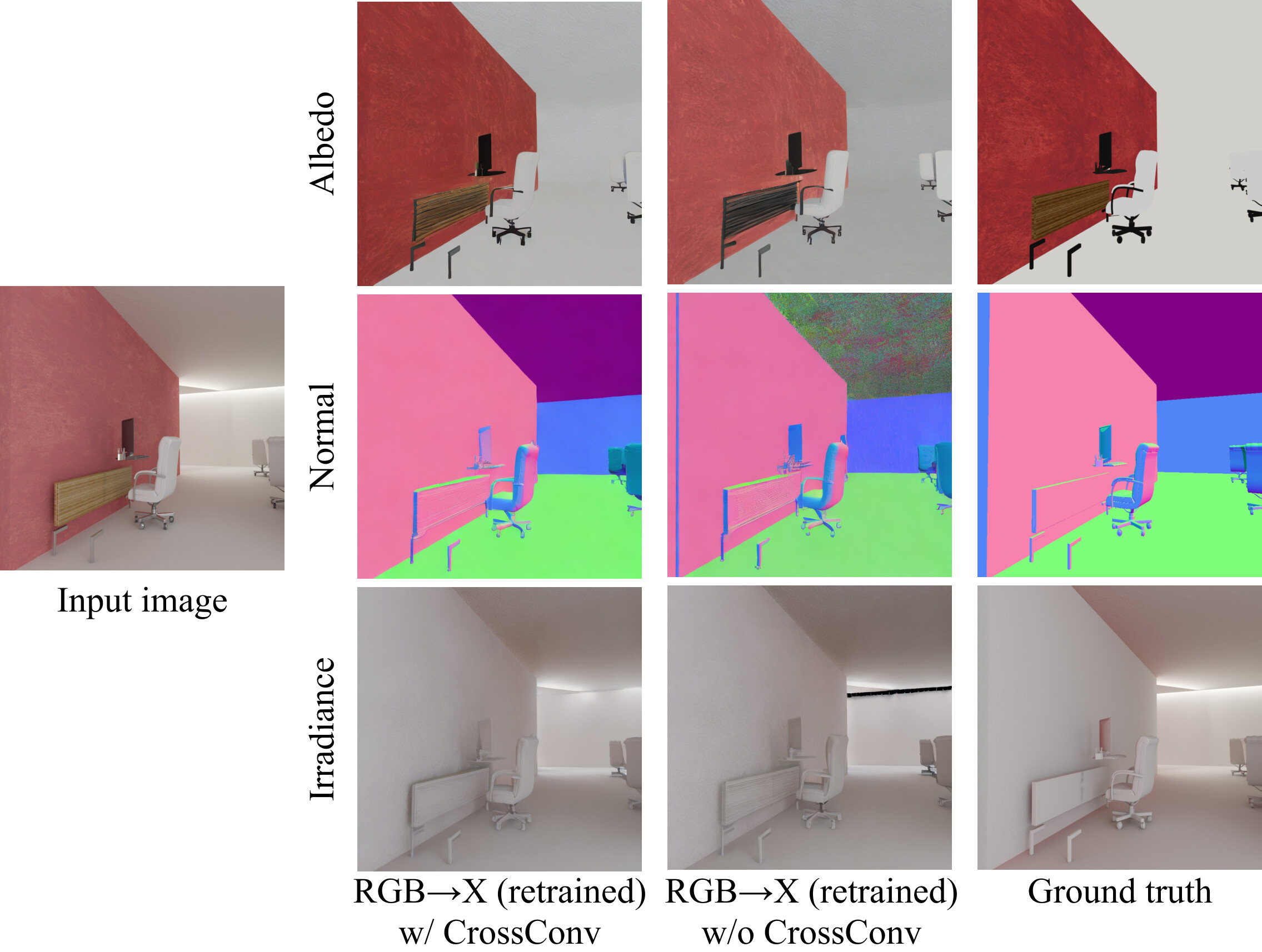}
    \caption{Effect of CrossStitch on intrinsic decomposition. By retraining RGB$\leftrightarrow$X~\cite{zeng2024rgb} on Hypersim~\cite{roberts2021hypersim}, we demonstrate that CrossStitch effectively improves structural consistency across decomposed outputs, even when applied to a U-Net denoising backbone, leading to better overall reconstruction.}
    \label{fig:abl_rgbx}
\end{figure}

\paragraph*{CrossStitch for Intrinsic Decomposition.} 
To evaluate the adaptability of CrossStitch beyond material generation, we apply it to the intrinsic decomposition task on the Hypersim dataset~\cite{roberts2021hypersim}, which contains over 74K indoor images with ground-truth albedo, normal, and irradiance maps. We \revised{retrained} RGB$\leftrightarrow$X~\cite{zeng2024rgb} based on Stable Diffusion 2.1~\cite{rombach2022LDM}, with and without our CrossStitch module. \revised{As shown in Tab.~\ref{tab:abl_rgbx} and Fig.~\ref{fig:abl_rgbx}, incorporating CrossStitch consistently improves performance in both PSNR, LPIPS, and DreamSim~\cite{fu2023dreamsim} by enabling structural consistency across predicted maps}. Despite using only the publicly available training split (unlike RGB$\leftrightarrow$X, which uses larger private datasets), our model demonstrates strong generalization, validating CrossStitch’s plug-and-play applicability across architectures (e.g., U-Net) and tasks beyond DiT-based material generation.

\begin{figure}[b]
    \centering
    \includegraphics[width=.8\linewidth]{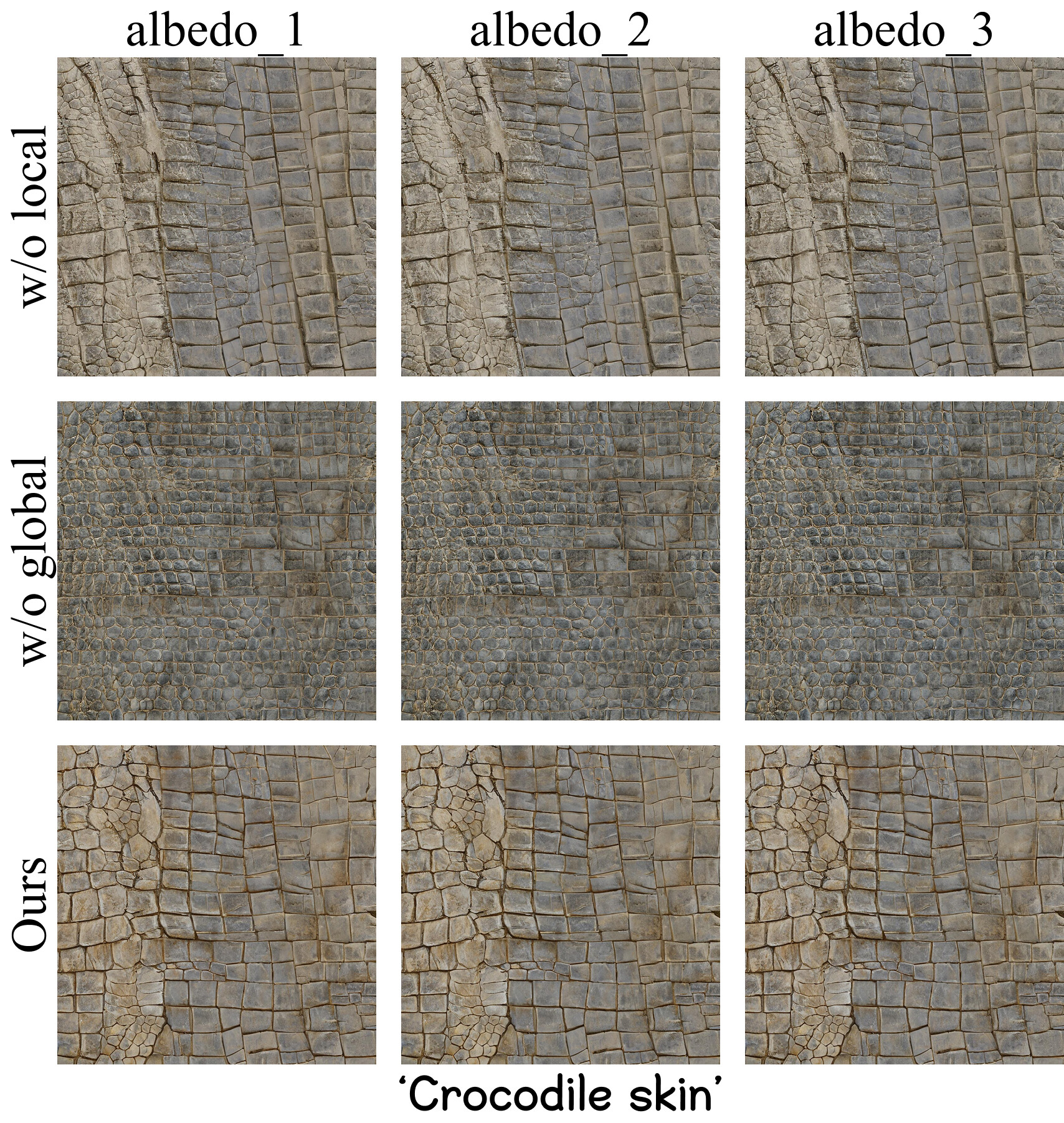}
    \caption{\revised{Visual results showing the effect of CrossStitch’s two branches. When both branches are enabled, the model produces more consistent results with reduced color-bias.}}
    \label{fig:abl_conv}
\end{figure}

\begin{table}[t]
\centering
\caption{\revised{Quantitative comparisons of different branch configurations within the CrossStitch module. Lower LPIPS and DreamSim scores indicate stronger perceptual consistency, highlighting the module’s potential to improve semantic alignment in multi-map generation tasks.}}
\label{tab:abl_conv}
\resizebox{\linewidth}{!}{%
\begin{tabular}{c|c|ccc} 
\toprule
                               & Pair & PSNR$\uparrow$ & LPIPS$\downarrow$ & DreamSim$\downarrow$  \\ 
\midrule
\multirow{4}{*}{Global branch} & 1$\leftrightarrow$2 & 33.13          & 0.07              & 0.0019                \\
                               & 1$\leftrightarrow$3 & 33.48          & 0.07              & 0.0019                \\
                               & 2$\leftrightarrow$3 & 32.32          & 0.07              & 0.0026                \\ 

\cmidrule{2-5}
                               & Mean                & 32.98          & 0.07              & 0.0021                \\ 
\midrule
\multirow{4}{*}{Local branch}  & 1$\leftrightarrow$2 & 32.01          & 0.06              & 0.0017                \\
                               & 1$\leftrightarrow$3 & 31.45          & 0.07              & 0.0038                \\
                               & 2$\leftrightarrow$3 & 31.79          & 0.06              & 0.0020                \\ 
\cmidrule{2-5}
                               & Mean                & 31.75          & 0.06              & 0.0025                \\ 

\midrule
\multirow{4}{*}{Both}          & 1$\leftrightarrow$2 & 34.85          & 0.06              & 0.0014                \\
                               & 1$\leftrightarrow$3 & 34.58          & 0.07              & 0.0016                \\
                               & 2$\leftrightarrow$3 & 34.59          & 0.07              & 0.0012                \\ 
\cmidrule{2-5}
                               & Mean                & \textbf{34.67} & \textbf{0.06}     & \textbf{0.0014}       \\
\bottomrule
\end{tabular}
}
\end{table}

\paragraph*{Effects of Different Branches in CrossStitch.}
\revised{
To analyze the effects of each component within the CrossStitch module, we conduct an ablation study by selectively turning its two branches on or off: (i) global branch: the average-pooling branch for global context aggregation, and (ii) local branch: the convolution branch for localized inter-map interactions. To quantitatively evaluate consistency, we set up the model to generate multiple output maps that are all intended to replicate the same albedo map, each associated with a different switcher name (e.g., albedo\_1, albedo\_2). Since all outputs originate from the same underlying signal, the ideal outputs should be identical, allowing us to directly assess consistency using image similarity metrics such as PSNR, LPIPS, and DreamSim~\cite{fu2023dreamsim}. From Tab.~\ref{tab:abl_conv} and Fig.~\ref{fig:abl_conv}, we observe that using either branch alone already enforces a reasonable level of consistency, while enabling both branches jointly more effectively suppresses color bias and yields the most consistent results across maps, demonstrating the full potential of the CrossStitch module for multi-map alignment.}

\begin{figure}[t]
    \centering  
    \includegraphics[width=\linewidth]{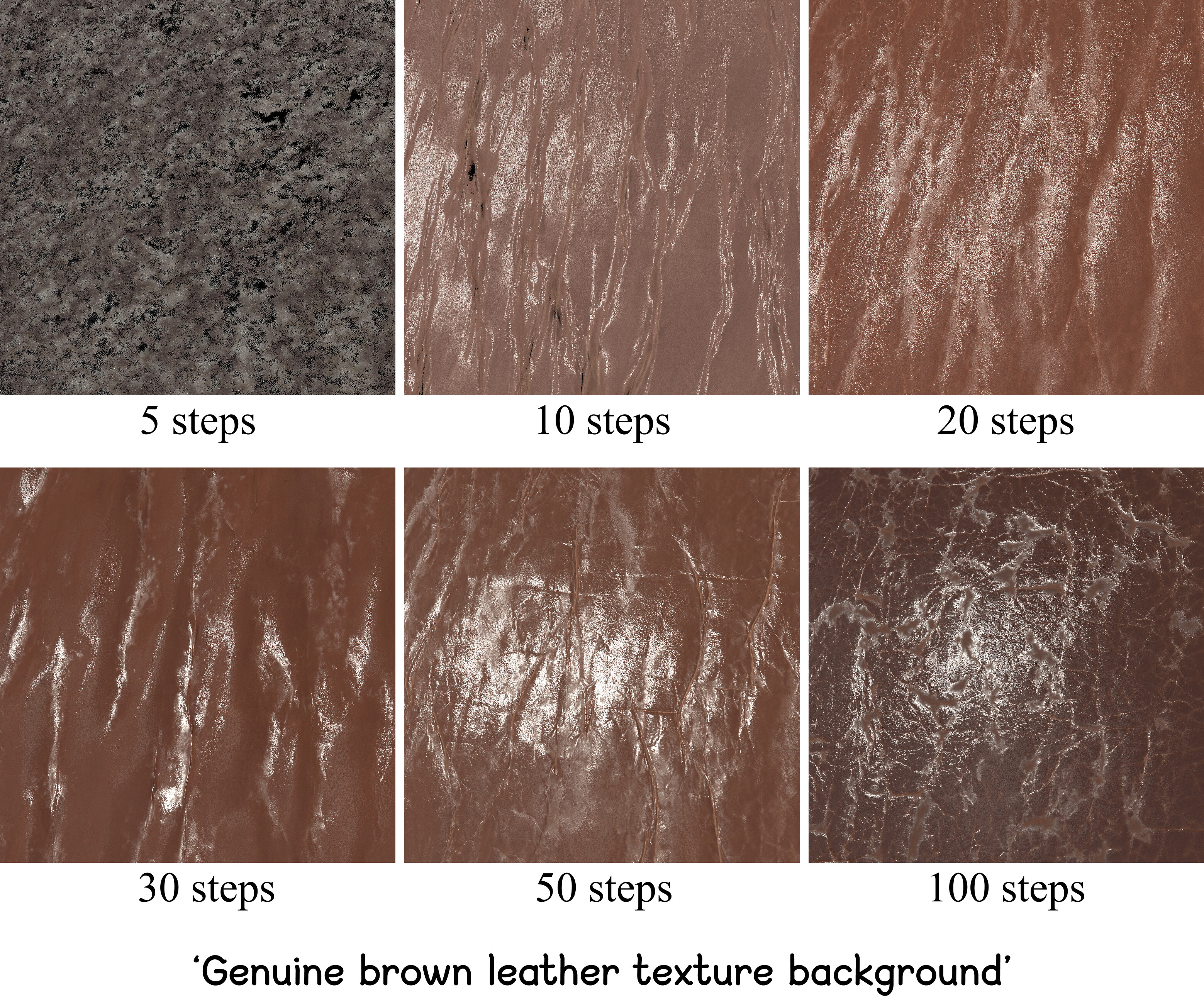}
    \caption{
    Visual comparison across different denoising steps. Increasing the number of sampling steps yields greater texture fidelity, including finer details.
    }
    \label{fig:abl_steps}
\end{figure}

\paragraph*{Effect of Denoising Steps.} We analyze the impact of the number of denoising steps during 4K-resolution generation. Specifically, we conduct experiments using 5, 10, 20, 30, 50, and 100 steps. As shown in Fig.~\ref{fig:abl_steps}, increasing the number of steps improves the preservation of fine details, but it also incurs a higher computational cost. By default, we adopt 20 steps to balance quality and efficiency, while allowing users to adjust this parameter according to application-specific requirements.




\subsection{Discussion and limitations}
Our method has several limitations. First, we observe that noise-rolling occasionally introduces horizontal and vertical streaks in 4K generation \revised{(see supplementary material for examples)}. Additional denoising stages can mitigate this artifact, though at the cost of increased inference time. Second, even with decoder fine-tuning, DC-AE could be further improved in preserving low-frequency structures~\cite{chen2025dcae1.5}. 
Finally, although our current implementation is text-conditioned, the framework readily generalizes to multimodal inputs (e.g., image prompts or ControlMat-style controls~\cite{vecchio2024controlmat}), offering greater flexibility and diversity, as with other diffusion-based methods. We leave this exploration for future work.


\section{Conclusion}
We propose \emph{HiMat}, a lightweight diffusion framework tailored for 4K SVBRDF generation. By combining a deep compression autoencoder with a linear-attention DiT, HiMat reduces the prohibitive pixel budget and scales generation efficiently to ultra-high resolution. To address the challenge of strict pixel alignment across reflectance maps, we introduce \emph{CrossStitch}, a convolutional module that enforces inter-map consistency while remaining non-destructive and compatible with standard diffusion backbones. Extensive experiments demonstrate that HiMat produces diverse, high-fidelity 4K SVBRDFs with practical runtime on consumer GPUs, establishing a new baseline for scalable material generation. Beyond materials, HiMat also generalizes naturally to related tasks such as intrinsic decomposition, highlighting its potential as a versatile foundation for efficient DiT-based pipelines in digital content creation.

\section*{Acknowledgments}
We thank the reviewers for the valuable comments. This work has been partially supported by the National Natural Science Foundation of China under grant No. 62572230.

\FloatBarrier
\bibliographystyle{eg-alpha-doi} 
\bibliography{bibliography}       

\newcommand{\etalchar}[1]{$^{#1}$}
\begin{thebibliography}{\uppercase{LCBH{\etalchar{*}}23}}

\bibitem[BPH{\etalchar{*}}24]{sora24}
\textsc{Brooks T., Peebles B., Holmes C., DePue W., Guo Y., Jing L., Schnurr D., Taylor J., Luhman T., Luhman E., Ng C., Wang R., Ramesh A.}:
\newblock Video generation models as world simulators.
\newblock URL: \url{https://openai.com/research/video-generation-models-as-world-simulators}.

\bibitem[CBS{\etalchar{*}}25]{comanici2025gemini}
\textsc{Comanici G., Bieber E., Schaekermann M., Pasupat I., Sachdeva N., Dhillon I., Blistein M., Ram O., Zhang D., Rosen E., et~al.}:
\newblock Gemini 2.5: Pushing the frontier with advanced reasoning, multimodality, long context, and next generation agentic capabilities.
\newblock \emph{arXiv preprint arXiv:2507.06261} (2025).

\bibitem[CCC{\etalchar{*}}25]{chen2025dcae}
\textsc{Chen J., Cai H., Chen J., Xie E., Yang S., Tang H., Li M., Han S.}:
\newblock Deep compression autoencoder for efficient high-resolution diffusion models.
\newblock In \emph{The Thirteenth International Conference on Learning Representations} (2025).

\bibitem[CGX{\etalchar{*}}24]{chen2024pixart}
\textsc{Chen J., Ge C., Xie E., Wu Y., Yao L., Ren X., Wang Z., Luo P., Lu H., Li Z.}:
\newblock Pixart-$\sigma$: Weak-to-strong training of diffusion transformer for 4k text-to-image generation.
\newblock In \emph{European Conference on Computer Vision} (2024), Springer, pp.~74--91.

\bibitem[CT82]{cook1982reflectance}
\textsc{Cook R.~L., Torrance K.~E.}:
\newblock A reflectance model for computer graphics.
\newblock \emph{ACM Transactions on Graphics (ToG) 1}, 1 (1982), 7--24.

\bibitem[CWH{\etalchar{*}}24]{chen2024dtdmat}
\textsc{Chen M., Wang Y., Hu D., Zhu P., Guo J., Guo Y.}:
\newblock Dtdmat: A comprehensive svbrdf dataset with detailed text descriptions.
\newblock In \emph{The 19th ACM SIGGRAPH International Conference on Virtual-Reality Continuum and its Applications in Industry} (2024), pp.~1--15.

\bibitem[CZH{\etalchar{*}}25]{chen2025dcae1.5}
\textsc{Chen J., Zou D., He W., Chen J., Xie E., Han S., Cai H.}:
\newblock Dc-ae 1.5: Accelerating diffusion model convergence with structured latent space.
\newblock In \emph{IEEE International Conference on Computer Vision (ICCV)} (2025).

\bibitem[DAD{\etalchar{*}}18]{deschaintre2018single}
\textsc{Deschaintre V., Aittala M., Durand F., Drettakis G., Bousseau A.}:
\newblock Single-image svbrdf capture with a rendering-aware deep network.
\newblock \emph{ACM Transactions on Graphics (ToG) 37}, 4 (2018), 1--15.

\bibitem[DAD{\etalchar{*}}19]{DADDB19}
\textsc{Deschaintre V., Aittala M., Durand F., Drettakis G., Bousseau A.}:
\newblock Flexible svbrdf capture with a multi-image deep network.
\newblock \emph{Computer Graphics Forum (Proceedings of the Eurographics Symposium on Rendering) 38}, 4 (July 2019).
\newblock URL: \url{http://www-sop.inria.fr/reves/Basilic/2019/DADDB19}.

\bibitem[DCH{\etalchar{*}}24]{du2024demofusion}
\textsc{Du R., Chang D., Hospedales T., Song Y.-Z., Ma Z.}:
\newblock Demofusion: Democratising high-resolution image generation with no \$\$\$.
\newblock In \emph{CVPR} (2024).

\bibitem[DN21]{dhariwal2021diffusion}
\textsc{Dhariwal P., Nichol A.}:
\newblock Diffusion models beat gans on image synthesis.
\newblock \emph{Advances in neural information processing systems 34} (2021), 8780--8794.

\bibitem[EKB{\etalchar{*}}24]{esser2024sd3}
\textsc{Esser P., Kulal S., Blattmann A., Entezari R., M{\"u}ller J., Saini H., Levi Y., Lorenz D., Sauer A., Boesel F., et~al.}:
\newblock Scaling rectified flow transformers for high-resolution image synthesis.
\newblock In \emph{Forty-first international conference on machine learning} (2024).

\bibitem[FTS{\etalchar{*}}23]{fu2023dreamsim}
\textsc{Fu S., Tamir N., Sundaram S., Chai L., Zhang R., Dekel T., Isola P.}:
\newblock Dreamsim: Learning new dimensions of human visual similarity using synthetic data.
\newblock In \emph{Advances in Neural Information Processing Systems} (2023), vol.~36, pp.~50742--50768.

\bibitem[GHS{\etalchar{*}}22]{guerrero2022matformer}
\textsc{Guerrero P., Hasan M., Sunkavalli K., Mech R., Boubekeur T., Mitra N.}:
\newblock Matformer: A generative model for procedural materials.
\newblock \emph{ACM Trans. Graph. 41}, 4 (2022).
\newblock \href {https://doi.org/10.1145/3528223.3530173} {\path{doi:10.1145/3528223.3530173}}.

\bibitem[GSH{\etalchar{*}}20]{Guo:2020:MaterialGAN}
\textsc{Guo Y., Smith C., Ha\v{s}an M., Sunkavalli K., Zhao S.}:
\newblock Materialgan: Reflectance capture using a generative svbrdf model.
\newblock \emph{ACM Trans. Graph. 39}, 6 (2020), 254:1--254:13.

\bibitem[HBG{\etalchar{*}}24]{hu2024zigma}
\textsc{Hu V.~T., Baumann S.~A., Gui M., Grebenkova O., Ma P., Schusterbauer J., Ommer B.}:
\newblock Zigma: A dit-style zigzag mamba diffusion model.
\newblock In \emph{ECCV} (2024).

\bibitem[HGH{\etalchar{*}}23]{hu2023gen}
\textsc{Hu Y., Guerrero P., Hasan M., Rushmeier H., Deschaintre V.}:
\newblock {Generating Procedural Materials from Text or Image Prompts}.
\newblock In \emph{ACM SIGGRAPH 2023 Conference Proceedings} (2023).

\bibitem[HGZ{\etalchar{*}}23]{he23text2mat}
\textsc{He Z., Guo J., Zhang Y., Tu Q., Chen M., Guo Y., Wang P., Dai W.}:
\newblock {Text2Mat: Generating Materials from Text}.
\newblock In \emph{Pacific Graphics Short Papers and Posters} (2023), The Eurographics Association.

\bibitem[HHF{\etalchar{*}}21]{hessel2021clipscore}
\textsc{Hessel J., Holtzman A., Forbes M., Bras R.~L., Choi Y.}:
\newblock Clipscore: A reference-free evaluation metric for image captioning.
\newblock \emph{arXiv preprint arXiv:2104.08718} (2021).

\bibitem[KHM{\etalchar{*}}24]{kavoosighafi2024deep}
\textsc{Kavoosighafi B., Hajisharif S., Miandji E., Baravdish G., Cao W., Unger J.}:
\newblock Deep svbrdf acquisition and modelling: A survey.
\newblock In \emph{Computer Graphics Forum} (2024), vol.~43, Wiley Online Library, p.~e15199.

\bibitem[KHZP25]{kim2025diffusehigh}
\textsc{Kim Y., Hwang G., Zhang J., Park E.}:
\newblock Diffusehigh: Training-free progressive high-resolution image synthesis through structure guidance.
\newblock In \emph{Proceedings of the AAAI conference on artificial intelligence} (2025), vol.~39, pp.~4338--4346.

\bibitem[KW14]{Kingma2014VAE}
\textsc{Kingma D.~P., Welling M.}:
\newblock {Auto-Encoding Variational Bayes}.
\newblock In \emph{The 2nd International Conference on Learning Representations} (2014).

\bibitem[Lab24]{flux2024}
\textsc{Labs B.~F.}:
\newblock Flux.
\newblock \url{https://github.com/black-forest-labs/flux}, 2024.

\bibitem[LCBH{\etalchar{*}}23]{lipman2023flow}
\textsc{Lipman Y., Chen R. T.~Q., Ben-Hamu H., Nickel M., Le M.}:
\newblock Flow matching for generative modeling.
\newblock In \emph{The Eleventh International Conference on Learning Representations} (2023).

\bibitem[LG16]{lavin2016fast}
\textsc{Lavin A., Gray S.}:
\newblock Fast algorithms for convolutional neural networks.
\newblock In \emph{Proceedings of the IEEE conference on computer vision and pattern recognition} (2016), pp.~4013--4021.

\bibitem[LLY{\etalchar{*}}21]{li2021convsurvey}
\textsc{Li Z., Liu F., Yang W., Peng S., Zhou J.}:
\newblock A survey of convolutional neural networks: analysis, applications, and prospects.
\newblock \emph{IEEE transactions on neural networks and learning systems 33}, 12 (2021), 6999--7019.

\bibitem[MCS23]{memery2023generating}
\textsc{Memery S., Cedron O., Subr K.}:
\newblock Generating parametric brdfs from natural language descriptions.
\newblock In \emph{Computer graphics forum} (2023), vol.~42, Wiley Online Library, p.~e14980.

\bibitem[MDH{\etalchar{*}}25]{ma2024materialpicker}
\textsc{Ma X., Deschaintre V., Ha{\v{s}}an M., Luan F., Zhou K., Wu H., Hu Y.}:
\newblock Materialpicker: Multi-modal material generation with diffusion transformers.
\newblock \emph{ACM Trans. Graph.} (July 2025).

\bibitem[MXZ{\etalchar{*}}23]{ma2023opensvbrdf}
\textsc{Ma X., Xu X., Zhang L., Zhou K., Wu H.}:
\newblock Opensvbrdf: a database of measured spatially-varying reflectance.
\newblock \emph{ACM Transactions on Graphics (TOG) 42}, 6 (2023), 1--14.

\bibitem[PDD{\etalchar{*}}24]{phung2024dimsum}
\textsc{Phung H., Dao Q., Dao T., Phan H., Metaxas D., Tran A.}:
\newblock Dimsum: Diffusion mamba - a scalable and unified spatial-frequency method for image generation.
\newblock In \emph{The Thirty-eighth Annual Conference on Neural Information Processing Systems} (2024).

\bibitem[PX23]{Peebles_2023DiT}
\textsc{Peebles W., Xie S.}:
\newblock Scalable diffusion models with transformers.
\newblock In \emph{Proceedings of the IEEE/CVF International Conference on Computer Vision (ICCV)} (October 2023), pp.~4195--4205.

\bibitem[QZX{\etalchar{*}}25]{lumina2}
\textsc{Qin Q., Zhuo L., Xin Y., Du R., Li Z., Fu B., Lu Y., Li X., Liu D., Zhu X., Beddow W., Millon E., Victor~Perez W.~W., Qiao Y., Zhang B., Liu X., Li H., Xu C., Gao P.}:
\newblock Lumina-image 2.0: A unified and efficient image generative framework, 2025.

\bibitem[RBL{\etalchar{*}}22]{rombach2022LDM}
\textsc{Rombach R., Blattmann A., Lorenz D., Esser P., Ommer B.}:
\newblock High-resolution image synthesis with latent diffusion models.
\newblock In \emph{Proceedings of the IEEE/CVF conference on computer vision and pattern recognition} (2022), pp.~10684--10695.

\bibitem[RDN{\etalchar{*}}22]{ramesh2022hierarchicaltextconditionalimagegeneration}
\textsc{Ramesh A., Dhariwal P., Nichol A., Chu C., Chen M.}:
\newblock Hierarchical text-conditional image generation with clip latents, 2022.
\newblock URL: \url{https://arxiv.org/abs/2204.06125}, \href {http://arxiv.org/abs/2204.06125} {\path{arXiv:2204.06125}}.

\bibitem[Rog22]{rogozhnikov2022einops}
\textsc{Rogozhnikov A.}:
\newblock Einops: Clear and reliable tensor manipulations with einstein-like notation.
\newblock In \emph{International Conference on Learning Representations} (2022).

\bibitem[RRR{\etalchar{*}}21]{roberts2021hypersim}
\textsc{Roberts M., Ramapuram J., Ranjan A., Kumar A., Bautista M.~A., Paczan N., Webb R., Susskind J.~M.}:
\newblock Hypersim: A photorealistic synthetic dataset for holistic indoor scene understanding.
\newblock In \emph{Proceedings of the IEEE/CVF international conference on computer vision} (2021), pp.~10912--10922.

\bibitem[SBV{\etalchar{*}}22]{schuhmann2022laion}
\textsc{Schuhmann C., Beaumont R., Vencu R., Gordon C., Wightman R., Cherti M., Coombes T., Katta A., Mullis C., Wortsman M., et~al.}:
\newblock Laion-5b: An open large-scale dataset for training next generation image-text models.
\newblock \emph{Advances in neural information processing systems 35} (2022), 25278--25294.

\bibitem[SP23]{matfusion}
\textsc{Sartor S., Peers P.}:
\newblock Matfusion: a generative diffusion model for svbrdf capture.
\newblock In \emph{ACM SIGGRAPH Asia Conference Proceedings} (December 2023).
\newblock URL: \url{https://doi.org/10.1145/3610548.3618194}.

\bibitem[SPN{\etalchar{*}}16]{schmidt2016state}
\textsc{Schmidt T.-W., Pellacini F., Nowrouzezahrai D., Jarosz W., Dachsbacher C.}:
\newblock State of the art in artistic editing of appearance, lighting and material.
\newblock In \emph{Computer Graphics Forum} (2016), vol.~35, Wiley Online Library, pp.~216--233.

\bibitem[TRP{\etalchar{*}}24]{team2024gemma}
\textsc{Team G., Riviere M., Pathak S., Sessa P.~G., Hardin C., Bhupatiraju S., Hussenot L., Mesnard T., Shahriari B., Ram{\'e} A., et~al.}:
\newblock Gemma 2: Improving open language models at a practical size.
\newblock \emph{arXiv preprint arXiv:2408.00118} (2024).

\bibitem[VD24]{vecchio2023matsynth}
\textsc{Vecchio G., Deschaintre V.}:
\newblock Matsynth: A modern pbr materials dataset.
\newblock In \emph{Proceedings of the IEEE/CVF Conference on Computer Vision and Pattern Recognition} (2024).

\bibitem[VMR{\etalchar{*}}24]{vecchio2024controlmat}
\textsc{Vecchio G., Martin R., Roullier A., Kaiser A., Rouffet R., Deschaintre V., Boubekeur T.}:
\newblock Controlmat: A controlled generative approach to material capture.
\newblock \emph{ACM Trans. Graph. 43}, 5 (sep 2024).
\newblock \href {https://doi.org/10.1145/3688830} {\path{doi:10.1145/3688830}}.

\bibitem[VSP{\etalchar{*}}17]{attentionall17}
\textsc{Vaswani A., Shazeer N., Parmar N., Uszkoreit J., Jones L., Gomez A.~N., Kaiser L., Polosukhin I.}:
\newblock Attention is all you need.
\newblock In \emph{Proceedings of the 31st International Conference on Neural Information Processing Systems} (Red Hook, NY, USA, 2017), NIPS'17, Curran Associates Inc., p.~6000–6010.

\bibitem[VSPS24]{vecchio2024matfuse}
\textsc{Vecchio G., Sortino R., Palazzo S., Spampinato C.}:
\newblock Matfuse: Controllable material generation with diffusion models.
\newblock In \emph{Proceedings of the IEEE/CVF Conference on Computer Vision and Pattern Recognition (CVPR)} (June 2024), pp.~4429--4438.

\bibitem[WMLT07]{walter2007microfacet}
\textsc{Walter B., Marschner S.~R., Li H., Torrance K.~E.}:
\newblock Microfacet models for refraction through rough surfaces.
\newblock \emph{Rendering techniques 2007} (2007), 18th.

\bibitem[WZZ{\etalchar{*}}24]{qalign}
\textsc{Wu H., Zhang Z., Zhang W., Chen C., Liao L., Li C., Gao Y., Wang A., Zhang E., Sun W., Yan Q., Min X., Zhai G., Lin W.}:
\newblock Q-align: teaching lmms for visual scoring via discrete text-defined levels.
\newblock In \emph{Proceedings of the 41st International Conference on Machine Learning} (2024), ICML'24, JMLR.org.

\bibitem[XCC{\etalchar{*}}25]{xie2025sana}
\textsc{Xie E., Chen J., Chen J., Cai H., Tang H., Lin Y., Zhang Z., Li M., Zhu L., Lu Y., Han S.}:
\newblock {SANA}: Efficient high-resolution text-to-image synthesis with linear diffusion transformers.
\newblock In \emph{The Thirteenth International Conference on Learning Representations} (2025).

\bibitem[XCZ{\etalchar{*}}25]{2025sana15}
\textsc{Xie E., Chen J., Zhao Y., Yu J., Zhu L., Lin Y., Zhang Z., Li M., Chen J., Cai H., Liu B., Zhou D., Han S.}:
\newblock Sana 1.5: Efficient scaling of training-time and inference-time compute in linear diffusion transformer.
\newblock In \emph{International Conference on Machine Learning} (January 2025).

\bibitem[XGZM24]{xue2024reflectancefusion}
\textsc{Xue B., Guarnera C., Zhao S., Montazeri Z.}:
\newblock Reflectancefusion: Diffusion-based text to svbrdf generation.
\newblock In \emph{Eurographics Symposium on Rendering} (2024), Eurographics Association.

\bibitem[XZW{\etalchar{*}}25]{xin2024dreampbr}
\textsc{Xin L., Zhang Z., Wei J., Gao W., Gao D.}:
\newblock Dreampbr: Text-driven generation of high-resolution svbrdf with multi-modal guidance.
\newblock \emph{IEEE International Conference on Multimedia \& Expo(ICME)} (2025).

\bibitem[YGL{\etalchar{*}}24]{yu2024scaling}
\textsc{Yu F., Gu J., Li Z., Hu J., Kong X., Wang X., He J., Qiao Y., Dong C.}:
\newblock Scaling up to excellence: Practicing model scaling for photo-realistic image restoration in the wild.
\newblock In \emph{Proceedings of the IEEE/CVF Conference on Computer Vision and Pattern Recognition} (2024), pp.~25669--25680.

\bibitem[YLTW25]{yu2025urae}
\textsc{Yu R., Liu S., Tan Z., Wang X.}:
\newblock Ultra-resolution adaptation with ease.
\newblock \emph{International Conference on Machine Learning} (2025).

\bibitem[YTZ{\etalchar{*}}25]{yang2024cogvideox}
\textsc{Yang Z., Teng J., Zheng W., Ding M., Huang S., Xu J., Yang Y., Hong W., Zhang X., Feng G., Yin D., Yuxuan.Zhang, Wang W., Cheng Y., Xu B., Gu X., Dong Y., Tang J.}:
\newblock Cogvideox: Text-to-video diffusion models with an expert transformer.
\newblock In \emph{The Thirteenth International Conference on Learning Representations} (2025).

\bibitem[YYSF24]{DiffMat24}
\textsc{Yuan L., Yan D., Saito S., Fujishiro I.}:
\newblock Diffmat: Latent diffusion models for image-guided material generation.
\newblock \emph{Visual Informatics 8}, 1 (2024), 6--14.

\bibitem[YZS{\etalchar{*}}23]{yang2023diffusion}
\textsc{Yang L., Zhang Z., Song Y., Hong S., Xu R., Zhao Y., Zhang W., Cui B., Yang M.-H.}:
\newblock Diffusion models: A comprehensive survey of methods and applications.
\newblock \emph{ACM Computing Surveys 56}, 4 (2023), 1--39.

\bibitem[ZDG{\etalchar{*}}24]{zeng2024rgb}
\textsc{Zeng Z., Deschaintre V., Georgiev I., Hold-Geoffroy Y., Hu Y., Luan F., Yan L.-Q., Ha\v{s}an M.}:
\newblock {RGB{$\leftrightarrow$}X}: Image decomposition and synthesis using material- and lighting-aware diffusion models.
\newblock In \emph{ACM SIGGRAPH 2024 Conference Papers} (New York, NY, USA, 2024), SIGGRAPH '24, Association for Computing Machinery.
\newblock \href {https://doi.org/10.1145/3641519.3657445} {\path{doi:10.1145/3641519.3657445}}.

\bibitem[ZHD{\etalchar{*}}22]{tilegen}
\textsc{Zhou X., Hasan M., Deschaintre V., Guerrero P., Sunkavalli K., Kalantari N.~K.}:
\newblock Tilegen: Tileable, controllable material generation and capture.
\newblock In \emph{SIGGRAPH Asia 2022 Conference Papers} (New York, NY, USA, 2022), SA '22, Association for Computing Machinery.

\bibitem[ZHD{\etalchar{*}}23]{zhou2023PhotoMat}
\textsc{Zhou X., Hašan M., Deschaintre V., Guerrero P., Hold-Geoffroy Y., Sunkavalli K., Kalantari N.~K.}:
\newblock Photomat: A material generator learned from single flash photos.
\newblock In \emph{SIGGRAPH 2023 Conference Papers} (2023).

\bibitem[ZHL{\etalchar{*}}25]{zhang2025diffusion4k}
\textsc{Zhang J., Huang Q., Liu J., Guo X., Huang D.}:
\newblock Diffusion-4k: Ultra-high-resolution image synthesis with latent diffusion models.
\newblock In \emph{IEEE/CVF Conference on Computer Vision and Pattern Recognition (CVPR)} (2025).

\bibitem[ZIE{\etalchar{*}}18]{zhang2018perceptual}
\textsc{Zhang R., Isola P., Efros A.~A., Shechtman E., Wang O.}:
\newblock The unreasonable effectiveness of deep features as a perceptual metric.
\newblock In \emph{CVPR} (2018).

\end{thebibliography}


\end{document}